\documentclass[11pt]{article}

\usepackage[letterpaper,margin=1in]{geometry}
\usepackage[T1]{fontenc}
\usepackage{lmodern}

\usepackage{microtype}
\usepackage{graphicx}
\usepackage{subcaption}
\usepackage{booktabs}
\usepackage{multirow}
\usepackage{xcolor}
\usepackage[round]{natbib}
\usepackage{caption}
\usepackage{hyperref}
\hypersetup{colorlinks=true,linkcolor=red!50!black,citecolor=blue!50!black,urlcolor=blue!60!black}
\usepackage{xurl}

\usepackage{amsmath}
\usepackage{amssymb}
\usepackage{mathtools}
\usepackage{amsthm}
\usepackage[capitalize,noabbrev]{cleveref}

\theoremstyle{definition}
\newtheorem{definition}{Definition}

\newcommand{\method}{WorldSketch}
\newcommand{\M}{\mathcal{M}}
\newcommand{\C}{\mathcal{C}}
\newcommand{\E}{\mathcal{E}}

\newif\ifshowtodos \showtodostrue

\title{\method{}: Sketching World Models for Zero-Shot Mobile Manipulation}
\author{Jiaming Wang}
\date{}

\begin{document}

\maketitle

\begin{abstract}
Most agentic robotic systems use vision-language models (VLMs) either to orchestrate predefined skills or to directly generate actions, constraints, and control programs. The former is limited by the available skill library, while the latter requires VLMs to reason about precise geometry and contact. In this paper, we propose \method{}, a new paradigm for agentic robotics in which a VLM constructs a world model rather than generating robot actions. The VLM acts as an agent that actively queries perception tools to build and visually validate a minimal kinematic world model, capturing only task-relevant objects, their interaction geometry, and possible motions. A geometric planner then jointly optimizes contact and whole-body motion on the resulting model. Moreover, the world model is self-improving: execution feedback verifies outcomes, updates uncertain hypotheses, and enables the agent to revise the model when its predictions are contradicted. We demonstrate the generality of this paradigm across RoboCasa, LIBERO-PRO, and R2R-CE navigation, achieving zero-shot success rates of 91\%, 97\%, and 76\%, respectively. On RoboCasa, \method{} outperforms the next-best zero-shot agent baseline by 58 percentage points while using 10$\times$ fewer VLM input tokens; on LIBERO-PRO, it exceeds CaP-X by 82 percentage points. In navigation, it matches the success rate of step-by-step VLM agents while using 6$\times$ fewer input tokens.

\end{abstract}

\section{Introduction}
\label{sec:intro}

Vision-language-action models (VLAs) map images and language to robot actions after training on large robot datasets \citep{zitkovich2023rt2,kim2024openvla,black2025pi0}.
However, they generalize poorly beyond the layouts and tasks they were trained on.
Policies that solve 82--99\% of the standard LIBERO suites, for example, succeed in 0--21\% of episodes on average once the initial object positions change, and in at most 1\% once the task changes \citep{zhou2025liberopro}.
Vision-language models (VLMs), trained on web-scale data, generalize much better across tasks, and they have become the decision maker of agentic robot systems that plan, call tools and improve themselves from feedback \citep{wang2026agenticrobotics}.

\begin{figure}[t]
  \centering
  \includegraphics[width=0.62\linewidth]{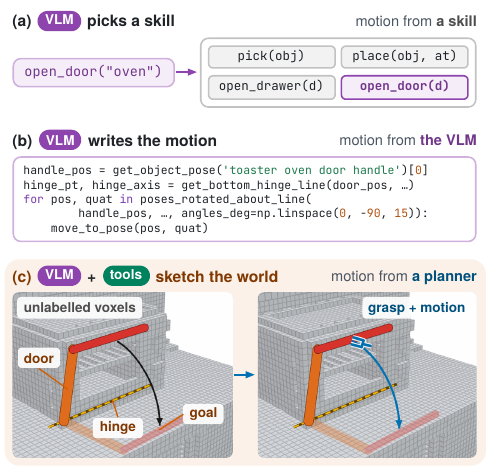}
  \caption{Three jobs for a VLM in a robot system, shown on the episode of \cref{fig:overview}. (a) The VLM picks the next skill from a library, and the skill moves the robot. (b) The VLM writes the motion itself, here the program that Code as Policies wrote for this episode. (c) In \method{}, the VLM sketches a minimal kinematic world model with perception tools. The door becomes a link on a hinge with a goal angle, the rest of the kitchen stays as unlabelled voxels, and a planner finds the grasp and a collision-free motion on this sketch.}
  \label{fig:teaser}
  \vspace{-2mm}
\end{figure}

In these agentic systems~\cite{wang2026agenticrobotics}, the VLM does one of two jobs (\cref{fig:teaser}a,b).
It either \emph{orchestrates} the task, choosing the next call from a library of skills \citep{ichter2022saycan,huang2022innermonologue}, or it \emph{writes the motion}, as value maps, keypoint constraints and waypoints \citep{huang2023voxposer,huang2024rekep,fang2024moka} or as control code that runs as the policy \citep{liang2023codeaspolicies,fu2026capx}.
Each job has a drawback.
An orchestrator is only as general as its skill library, while a VLM that writes motion or code either stays inside the slow control loop or must reason directly about metric geometry and contact.

\begin{figure*}[t]
  \centering
  \includegraphics[width=\textwidth]{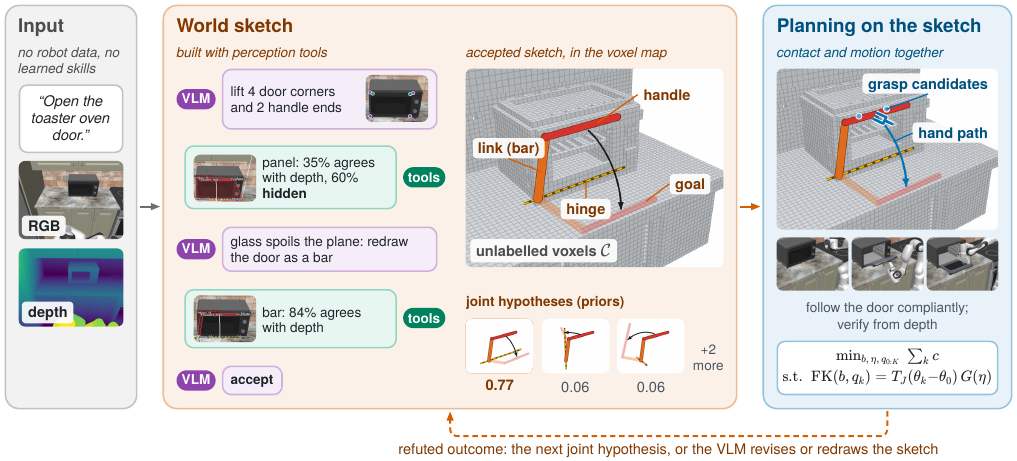}
  \caption{\method{} on a RoboCasa episode (\emph{open the toaster oven door}). The VLM builds the sketch in a short conversation with perception tools and checks it in rendered overlays, here redrawing the glass door as a bar after its first panel failed the depth test. The sketch holds only the door, with its handle, hinge, goal angle and ranked joint hypotheses, and the rest of the kitchen stays as voxels. The planner optimizes the grasp, the base placement and the arm motion along the door's arc together, and a refuted outcome leads to the next hypothesis, a revision or a fresh sketch.}
  \label{fig:overview}
  \vspace{-3mm}
\end{figure*}

In this paper, we give the VLM a third job, sketching the world (\cref{fig:teaser}c).
Instead of asking the VLM what the robot should do, we ask it what the robot needs to know about the world, and we leave the doing to a planner.
We ask the VLM to draw a \emph{world sketch}, a minimal world model of the task.
Like a sketch, it holds only what matters.
Only the task-relevant entities carry semantic labels.
To open a cabinet, for example, the sketch holds the door panel, its handle and its hinge, while all other objects in the kitchen remain unlabelled collision geometry.
The sketch is also kinematic.
The door is a link on a revolute joint with an axis, an origin and a goal angle, so that a planner can move the gripper along the door's arc.
Such a sketch is small enough for a VLM to draw from a handful of measurements, yet complete enough for a planner to act on.

We treat drawing the sketch as an agentic task (\cref{fig:overview}).
The VLM requests targeted perception, such as ``the 3-D points of the door's corners and of the handle's ends'' or ``a straight-on view of the stove's knob labels''.
It assembles a sketch from the measured primitives, sees this sketch rendered back into the images and corrects it.
Perception thus becomes a set of tools that the VLM calls, instead of a segmentation of the whole scene.
Properties that the images cannot settle, such as which way a faucet lever turns on the water, remain in the sketch as joint hypotheses with prior probabilities.
On the sketch, a planner chooses the grasp jointly with the whole-body motion along the joint, and a verifier confirms each outcome from changes in depth and appearance.
When execution contradicts the sketch, the robot tries the next hypothesis or the VLM revises the sketch, and after a failed attempt a fresh VLM session draws it anew.

We evaluate \method{} on 30 RoboCasa kitchen tasks with a mobile manipulator, where it succeeds in 91\% of episodes with about five VLM calls per episode and without any robot data.
With the same VLM, the CaP-X coding agent succeeds in 33\% and the other zero-shot baselines in at most 9\%, while two VLAs trained on demonstrations of these tasks succeed in 39\% and 44\%.
Success changes little with the VLM or its reasoning effort (86--91\%).
The first sketch is correct in 76--81\% of episodes, and in the runs we analyzed most failures happen in execution.
On LIBERO-PRO, whose tabletop tasks perturb object positions and instructions, \method{} succeeds in 97\% of episodes with about three VLM calls each, compared with 15\% for the CaP-X coding agent.
The idea also carries over to instruction-following navigation, where a sketch of the doors, passages and stairs that the instruction mentions guides a path planner to the goal.
On R2R-CE, \method{} succeeds as often as agents that ask the same VLM for each action (76\%), with a fifth of the VLM calls.

We make three contributions:
\begin{itemize}
  \setlength\itemsep{1pt}
  \item The world sketch, a minimal kinematic world model with joint hypotheses, as the interface between a VLM and a planner (\cref{sec:problem}).
  \item \method{}, an agentic system in which the VLM draws the sketch by requesting perception and checking the result, and a planner optimizes contact and motion on it. Execution evidence tests the sketch and triggers revisions (\cref{sec:method}).
  \item An evaluation on RoboCasa kitchen tasks against code-as-policies agents, keypoint constraints and two VLAs, on LIBERO-PRO tabletop tasks and on navigation, with analyses of the accuracy of the sketches and of where success is lost (\cref{sec:exp}).
\end{itemize}

\section{Related Work}
\label{sec:related}

\textbf{Agentic robot systems.}
Foundation models act as the decision maker of agentic robot systems in two main roles \citep{wang2026agenticrobotics}.
In \emph{plan generation}, the model decomposes an instruction into calls to a library of skills \citep{ichter2022saycan,huang2022innermonologue,singh2023progprompt}, or into goals and specifications for a task-and-motion planner whose domain model the designer supplies \citep{lin2023text2motion,ding2023llmgrop,chen2024autotamp,garrett2021tamp}.
In \emph{action generation}, the model produces the motion itself.
It writes control code that runs as the policy, either once or revised over turns from execution feedback \citep{liang2023codeaspolicies,zhi2025comerobot,fu2026capx}, writes value maps, keypoint constraints, marked waypoints or part poses that an optimizer turns into motion \citep{huang2023voxposer,huang2024rekep,fang2024moka,huang2024copa,pan2025omnimanip}, or chooses among sampled actions \citep{nasiriany2024pivot}.
Navigation agents likewise choose viewpoints, frontiers or pixels \citep{zhou2024navgpt,yokoyama2024vlfm,long2024instructnav}.

Self-improving systems also learn from execution feedback.
They explain failures \citep{liu2023reflect}, monitor constraints during execution \citep{zhou2025codeasmonitor}, revise their code over many turns \citep{fu2026capx}, or debug and search over programs on the target tasks, scored by the reward or the success check of the simulator \citep{elmaaroufi2026rho,lu2026aspire}.
Concurrent work drives robots with general-purpose coding agents through a terminal, a Python interpreter or a visual interface \citep{chu2026openrua,si2026pyrualean,hu2026via}.
In all of these roles, the output of the model is an action or a specification of one.
Metric geometry and contact are therefore decided inside the code or the constraints that the model writes, and what the robot finds out while acting is kept in a program, not in a model of the scene.
In contrast, we ask the VLM for this model and leave the actions to a planner.
Because the sketch is explicit, it can be rendered into images, tested against depth and corrected by interaction.

\textbf{VLMs that build models.}
Open-vocabulary maps and scene graphs give LLM planners a queryable world \citep{huang2023vlmaps,gu2024conceptgraphs,rana2023sayplan,jiang2024roboexp}, and LLMs can write symbolic planning domains \citep{liu2023llmp,guan2023worldmodels}, including hypotheses that execution tests \citep{xiao2026hume}.
These models are built for any task, and they are semantic or symbolic rather than metric and kinematic.
Articulated digital twins can be reconstructed from images as URDF or code, either by learned predictors or by VLMs with a render-and-critique loop \citep{chen2024urdformer,mandi2025real2code,le2025articulateanything}.
Digital cousins and twins are also built for policy training \citep{dai2024acdc,torne2024rialto}, and templates of articulated parts can be filled in by a regressor trained on part datasets \citep{sun2026grace}.
Interactive perception estimates articulation from the motion of the robot itself, using probabilistic models over joint types and parameters \citep{sturm2011probabilistic,martinmartin2014online,hausman2015active,bohg2017interactive}, and mobile manipulators have planned on such estimated models \citep{hsu2024kinscene,gupta2025mosart}.
These approaches aim at complete replicas of objects or scenes, or they rely on learned perception of all articulated parts.
In \method{}, we instead sketch online only the part of the world that the instruction needs, at the accuracy that the planner needs.
The sketch takes its hypotheses and priors from a VLM and lets the actions of the robot test them.

\textbf{VLAs and world models.}
VLAs learn actions end to end from large robot datasets \citep{zitkovich2023rt2,kim2024openvla,black2025pi0,black2025pi05,bjorck2025gr00t}.
They handle contact where the data cover it, but they transfer poorly to new layouts and tasks \citep{zhou2025liberopro}.
Learned world models predict future observations or latent states from actions, which can be used for planning or for training policies in imagination \citep{ha2018worldmodels,hafner2025dreamerv3,wu2022daydreamer,hansen2024tdmpc2,zhou2025dinowm}, and video models simulate interaction from pixels \citep{yang2024unisim,bruce2024genie}.
Both learn dense dynamics from data.
A world sketch lies at the other end of this spectrum.
It is an explicit model of the few bodies that the task acts on, built zero-shot by a VLM from measurements, with dynamics given by kinematics instead of learned from data, and it is accurate only where the planner needs it.

\section{World Sketches}
\label{sec:problem}

\textbf{Setting.}
A robot with known kinematics and calibrated RGB-D cameras receives a language instruction $\ell$ in a scene it has never seen.
It may move its base, torso and arm, and it observes its own joint state and wrist force.
We assume no demonstrations, no learned skills and no access to the simulator state, and we use ground truth only to score episodes.

\begin{definition}[World sketch]
A world sketch for instruction $\ell$ is a minimal world model $\M = (\C, \E, g)$ with three components.
\begin{itemize}
  \setlength\itemsep{0pt}
  \item $\C$ is the collision geometry, a set of voxels lifted from depth with no labels.
  \item $\E$ is the set of task entities, the parts that the task acts on, each with semantics, measured geometry and kinematics.
  \item $g$ is the goal, a target state of the entities.
\end{itemize}
\end{definition}

An articulated entity $e \in \E$ is a one-joint kinematic chain
\[
\text{world} \xrightarrow{\;J=(\tau,\,\mathbf{a},\,\mathbf{o},\,q)\;} \text{link } L \xrightarrow{\;\text{rigid}\;} \text{part } P .
\]
The joint $J$ has a type $\tau \in \{\text{prismatic}, \text{revolute}\}$, a unit axis $\mathbf{a}$, an origin $\mathbf{o}$ and a state $q$.
The link $L$ is the moving body, such as a door panel, a drawer front or a lever, and it is made of primitives such as panels, boxes and bars.
The part $P$ is where the robot acts on the link, for example a cylinder for a bar handle, a disk for a knob, or a button.
When an entity has no part, the robot pushes the link itself.
A point $\mathbf{x}$ on the link moves with the joint as $\mathbf{x}(q) = T_J(q - q_0)\,\mathbf{x}(q_0)$, where $T_J(\delta)$ is a rotation by $\delta$ about the line $(\mathbf{a},\mathbf{o})$ or a translation by $\delta\mathbf{a}$.
The goal for $e$ is a target value $q^\star$, possibly ``until the stop'', and optionally an effect region that shows that the task was done, such as the burner that a knob lights or the basin that a faucet fills.
A pick-and-place entity is an object primitive with a place region.

\textbf{Joint hypotheses.}
Images often do not determine the mechanism.
A faucet lever may turn the water on by tilting sideways or by lifting, and a knob may turn either way.
Each entity therefore carries a set of hypotheses $\{(J_k, p_k)\}_{k=1}^{K}$ with probabilities $p_k$.
The VLM proposes the hypotheses and their priors, and we add the common alternatives with a fraction of the prior, for example the opposite sense of a guessed direction, the other axes of a lever, or a press for a small part.
The sketch is thus a belief over mechanisms, which the actions of the robot test.

We call the sketch minimal because only the entities in $\E$ carry semantics and kinematics.
Everything else in the scene, such as the toaster on the counter, the neighbouring doors or the wall, matters to the task only as an obstacle, and $\C$ represents it as one.
We call a sketch \emph{sufficient} if a plan computed on it achieves the goal in the real scene, and we aim for the smallest sufficient sketch, in which an entity, a part or a joint parameter appears only when it changes what the planner will do.
Minimality follows from the construction procedure, because the VLM requests evidence only for the parts it needs and the backend measures only the requested quantities (\cref{sec:construction}).
In kitchens with over a thousand geometric bodies, the median sketch has one entity and two primitives (\cref{sec:exp_sketch}).

We ask the VLM for a sketch instead of actions because the sketch splits the work by competence.
The VLM decides which parts matter, which type of mechanism they have and what the goal is, a decision that requires common-sense and visual knowledge.
Depth measurements supply all metric quantities.
The planner computes grasps, collision-free whole-body motions and the path of the gripper along the joint, a computation that requires geometry and kinematics.
Because the sketch is explicit, it can also be drawn into images and checked by the VLM, tested against depth and corrected by interaction.

\section{\method{}}
\label{sec:method}

\Cref{fig:overview} shows the system.
We first describe how a VLM agent builds the sketch with perception tools and corrects it from execution evidence (\cref{sec:construction}), and then how a planner optimizes contact and motion on the sketch (\cref{sec:planning}).
In \cref{sec:domains}, we describe the instances for tabletop manipulation and navigation.

\subsection{Agentic Sketch Construction}
\label{sec:construction}

\begin{figure*}[t]
  \centering
  \includegraphics[width=\textwidth]{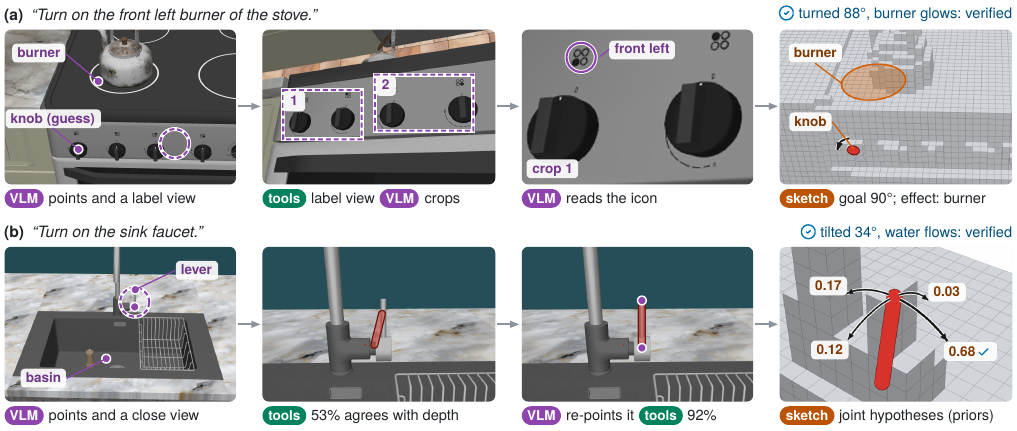}
  \caption{Sketch construction on two RoboCasa episodes, with the points that the VLM marks (purple dots), the views (dashed circles) and crops (dashed boxes) that it requests, and the accepted sketch in the voxel map of the planner (last image). (a) To find the knob of the front left burner, the VLM reads the burner icon in a label view and sketches the knob with the burner as its effect region. (b) The lever drawn from the first points agrees with depth in 53\% of its pixels, and in 92\% after the VLM re-points it. Of its four joint hypotheses, the most probable one (ticked) turned the water on. \Cref{app:tools} gives the transcripts.}
  \label{fig:construction}
  \vspace{-3mm}
\end{figure*}

The VLM receives the instruction and the images of the head and wrist cameras, and it builds the sketch over a few turns of one session (\cref{fig:construction}).
In each turn, it can request evidence, assemble or edit the sketch, and accept it (\cref{app:tools} lists the tools and gives full examples).

A request names what to measure, not how to segment the scene.
First, the VLM can ask for \emph{points}, which it marks in an image together with their kind, such as a corner, a surface point or a part like a knob or the end of a handle, and the backend lifts them to 3-D with depth.
Second, it can ask for \emph{views}, such as a new image of the wrist camera at a chosen pixel and distance, a crop of an earlier image, or a straight-on ``label view'' of a control panel, from which it reads the icons of knobs and buttons.
We ground requests in points rather than masks, because a small part such as a knob or a button is a single snapped point, which a segmentation mask often misses (\cref{sec:exp_ablation}).
The backend fits each point according to its kind, for example to the plane around a surface point or to the cluster that protrudes around a part point.
It also reports the result of each fit, or the reason why it failed, so that the VLM can recognize a poor measurement.

The VLM then writes the sketch as structured text.
It names the structure: which primitives form the link and the part, the joint type, the axis by keyword or by two points (``the left edge of the door''), the goal (``open until the stop'', ``turn to high''), joint hypotheses with priors, and an optional effect region.
It never writes a metric number that a measurement can supply, because the backend computes axes, origins and sizes from the lifted points.
After each assembly, we render the sketch into every image, with its joint axis, its goal pose and the path of the grasped point, and a depth test marks each primitive as agreeing with depth, floating in free space or hidden.
The VLM sees these overlays in its next turn and either revises the sketch or accepts it.
We found this check to matter most for the accuracy of the sketch (\cref{sec:exp_ablation}).

\textbf{Correcting the sketch from execution.}
The sketch is a belief that execution tests.
After each goal, a verifier compares images of the head cameras taken before and after the motion.
The goal counts as verified only when depth shows the link at the claimed joint value or when the effect region, such as a burner or a basin, changes (\cref{app:details}).
We do not let the joint reading of the gripper verify a goal, because the gripper may move without the part, for example after a slip.

The outcome $o$ of an attempt (verified, moved without effect, resisted, moved along another joint, and so on) updates the hypotheses by Bayes' rule,
\begin{equation}
  p_k \;\propto\; p_k \; P\big(o \,\big|\, \rho(k, j)\big),
\end{equation}
where $j$ is the executed hypothesis and $\rho(k,j) \in \{\text{same}, \text{flipped}, \text{other}\}$ relates hypothesis $k$ to it.
For example, an outcome of ``resisted'' makes the opposite sense likely.
The robot then executes the most probable hypothesis.
When no hypothesis of a part is still plausible, the VLM revises the sketch in the same session from the evidence of every attempt.
When an attempt fails as a whole, the robot returns to its initial pose and a fresh VLM session draws a new sketch, given a summary of the failed attempt.
A fresh session also re-examines successes that rest on weak evidence.

\subsection{Planning on the Sketch}
\label{sec:planning}

The planner treats every entity in the same way, without per-task skills.
It chooses where to touch the part together with the whole-body motion that follows the part to its goal.
We plan the two together because a contact chosen first often cannot be followed, for example when the arm reaches a joint limit or the palm meets the counter halfway along the arc of a door.

The planner samples contacts from the primitive of the part, such as grasps across a bar or a knob, presses on a button, and pushes on the faces of the link wherever the motion of the joint points into them (\cref{app:planner}).
A grasp at hand pose $H$ on a link at state $q_{\text{now}}$ moves rigidly with the link, $H(q) = T_J(q - q_{\text{now}})\,H$ for $q$ from $q_{\text{now}}$ to $q^\star$.
For each contact, the planner screens placements of the base and the torso by batched inverse kinematics along this hand path, and it selects the contact and placement for which the arm follows the path furthest without collision.
Finally, approach paths from the tucked arm come from RRT-Connect \citep{kuffner2000rrtconnect} in VAMP \citep{thomason2024vamp}, and objects to pick are grasped across their narrow side and released where the hand and the object have a free way in.

\textbf{Contact--motion optimization.}
A bar handle offers a continuous family of grasps, and the arc of a door leaves little room between the hand, the door and its frame.
For bars, the planner therefore optimizes the grasp, the placement and the arm motion jointly.
The grasp $G(\eta)$ has parameters $\eta = (s, \psi, d, w)$: the offset along the bar, the approach tilt about it, the grip depth and the finger opening.
With the base and torso displacement $b$ and arm configurations $q_k$ at joint samples $\theta_0 = q_{\text{now}}, \dots, \theta_K = q^\star$, it solves
\begin{equation}
\begin{aligned}
  \min_{b,\,\eta,\,q_{0:K}} \;& \textstyle\sum_k c_q(q_k, q_{k+1}) + c_b(b) + c_\eta(\eta) \\
  \text{s.t.} \;\;& \mathrm{FK}(b, q_k) = T_J(\theta_k - \theta_0)\,G(\eta), \\
  & h(\eta) \ge m_h, \;\; d(b, q_k, \theta_k) \ge m_a \;\;\; \forall k,
\end{aligned}
\end{equation}
where $\mathrm{FK}$ is the hand pose, $c_q$ penalizes joint motion, unnatural postures, joint-limit proximity and overstretched reach, $c_b$ penalizes base motion, and $c_\eta$ penalizes tilted, off-centre or shallow grasps.
The equality closes the kinematic chain, so that at every sample the hand holds the bar where the joint has moved it.
Because the hand is rigid on the bar, its clearance $h$ from the moving link, the posts of the bar and $\C$ depends only on $\eta$, from the approach to the release, and $d$ is the clearance of the arm and the base from $\C$ and from the link at $\theta_k$.

We solve this problem with an augmented-Lagrangian method from a few seeds and take the cheapest feasible solution as the plan.
When no solution covers the whole motion, the planner solves a prefix and re-plans the rest after execution (\cref{app:planner}).

\textbf{Closed-loop execution.}
We control the robot with compliant operational-space control \citep{khatib1987osc}, so that it follows the link.
It moves in small increments from the measured hand pose instead of along the planned arc, so that an error of a few centimetres in the hinge does not accumulate.
After every increment, it measures the joint value from the motion of the hand and refits the hinge to the path of the hand.
After each round, it re-observes the scene, corrects the joint value from depth when the view contradicts the hand, and re-plans with another grasp, a push or another base pose until the goal or a mechanical stop is reached.

\subsection{Other Domains}
\label{sec:domains}

\textbf{Tabletop manipulation.}
On LIBERO tabletops, every object is a separate cluster of depth points above the table, so we build the sketch by selection.
The backend proposes the clusters as numbered boxes, and the VLM promotes the task-relevant ones to entities, adds place regions and marks where to grasp, place, pull or turn.
The other clusters remain collision boxes.
A sampling-based planner computes the motions, depth and gripper evidence verify each event, and the VLM is called again when execution stops.

\textbf{Navigation.}
The same principle applies to following route instructions such as ``exit the room with the fireplace, turn right and go down the stairs''.
Here the collision geometry $\C$ is a 2.5-D occupancy grid built from depth, with no labels.
The task entities $\E$ are 3-D primitives that the VLM places from lifted points or map coordinates, such as doors, passages, staircases joining two floors, rooms, landmark objects and the goal, and each entity is checked against depth as the robot moves.
These entities change the cost maps of the planner, for example by opening the aperture of a door or making a staircase traversable.
In each turn, the VLM can look in a chosen direction, lift points, edit the sketch and send the robot towards a primitive, and a least-cost path planner and a controller execute the motion.

\section{Experiments}
\label{sec:exp}

In this section, we ask three questions.
How does sketching compare with code as policies, keypoint constraints and trained VLAs (\cref{sec:exp_main})?
How accurate are the sketches, and what makes them accurate (\cref{sec:exp_sketch})?
Where do the methods fail (\cref{sec:exp_failures})?
We also apply the idea to navigation (\cref{sec:exp_nav}).

\begin{table*}[t]
  \caption{RoboCasa kitchen tasks: successes per family (of 15) and in all (of 90), and per episode the VLM calls, the input and output tokens (with reasoning) and the minutes, with each VLM call counted by its response time (\cref{app:runtime}). \emph{False}: episodes declared solved that were not. The zero-shot methods use GPT-6.1-Sol, and the last row is the better of two runs with GPT-6-Astra (\cref{tab:vlm_final}). The time of ReKep is a lower bound, since 52 of its episodes reach a time limit.}
  \label{tab:main}
  \centering
  \small
  \setlength{\tabcolsep}{4pt}
  \resizebox{\textwidth}{!}{\begin{tabular}{lcccccccccccc}
    \toprule
    method & drawers & doors & knobs & buttons & levers & pick \& & all & calls & input & output & time & false \\
           & \& racks &      &       &         & \& spout & place &     &       & tokens & tokens & (min) & \\
    \midrule
    \multicolumn{13}{l}{\emph{trained on demonstrations of the tasks}} \\
    GR00T N1.5, 3 runs \citep{bjorck2025gr00t} & 8.7 & 8.3 & 6.0 & 4.7 & 8.0 & 4.3 & 40.0 (44\%) & -- & -- & -- & 0.8 & -- \\
    $\pi_{0.5}$ \citep{black2025pi05} & 10 & 5 & 2 & 1 & 8 & 9 & 35 (39\%) & -- & -- & -- & 1.2 & -- \\
    \midrule
    \multicolumn{13}{l}{\emph{zero-shot, GPT-6.1-Sol}} \\
    Code as Policies \citep{liang2023codeaspolicies} & 3 & 0 & 2 & 2 & 1 & 0 & 8 (9\%) & 6.1 & 31k & 1.5k & 1.9 & -- \\
    CaP-Agent0 (CaP-X) \citep{fu2026capx} & 6 & 2 & 7 & 12 & 0 & 3 & 30 (33\%) & 81 & 872k & 43k & 17.0 & 7 \\
    ReKep \citep{huang2024rekep} & 2 & 0 & 1 & 0 & 1 & 1 & 5 (6\%) & 1.0 & 6k & 1.0k & $\geq$97 & -- \\
    \method{} & 14 & 14 & 14 & 15 & 14 & 11 & \textbf{82} (91\%) & 5.4 & 88k & 1.8k & 7.1 & 2 \\
    \midrule
    \method{}, GPT-6-Astra & 15 & 14 & 15 & 13 & 14 & 11 & 82 (91\%) & 5.5 & 90k & 1.4k & 6.2 & 3 \\
    \bottomrule
  \end{tabular}}
  \vspace{-3mm}
\end{table*}

\subsection{Setup}
\label{sec:exp_setup}

\textbf{Kitchen tasks.}
We use RoboCasa365 \citep{nasiriany2026robocasa365} in MuJoCo \citep{todorov2012mujoco} with the PandaOmron mobile manipulator, a Panda arm on an omnidirectional base with a torso lift, three base-mounted RGB-D cameras and a wrist camera.
We evaluate 30 atomic tasks in six families of five (drawers and racks, doors, knobs, buttons, levers and spout, pick and place), with three episodes per task in the default kitchens of RoboCasa365, which gives 90 episodes in all.
Each family holds common tasks, such as opening a drawer or turning on a stove, and tasks drawn at random from the RoboCasa365 atomic tasks that act on the same kind of part (\cref{app:bench}).
We score every method with the success check of RoboCasa at the end of the episode, with one exception for the zero-shot methods.
RoboCasa also requires the gripper to end 15\,cm away from a pressed microwave or coffee-machine button, which the instruction does not state, so for these methods we count the press once the appliance has switched.
This rule changes five episodes of CaP-Agent0 and none of the other zero-shot methods, and the VLAs keep the full check of their evaluation protocol.

\textbf{Tabletop tasks.}
On LIBERO-PRO \citep{zhou2025liberopro}, which perturbs LIBERO tasks \citep{liu2023libero}, we use the position and task perturbations of the object, goal and spatial suites, with five randomly drawn tasks per suite and five initial states per task, which gives 150 episodes.
Success is the goal predicate of LIBERO in the final state.

\textbf{Baselines.}
We compare with three published zero-shot methods, which we ported from their released code to our robot while keeping their prompting and control logic (\cref{app:baselines}).
\emph{Code as Policies} (CaP; \citealp{liang2023codeaspolicies}) writes one program per instruction from few-shot examples and recursively generates the helper functions that the program calls.
It sees the scene only through the values that its perception functions return, and it gets no feedback.
\emph{CaP-Agent0}, the training-free agent of CaP-X \citep{fu2026capx}, writes a program and then sees its output and errors, together with a text description by a VLM of how the scene changed.
It then rewrites the program or finishes, for up to ten rewrites, and in each turn it merges three candidate programs.
Both methods call the same functions, which we modelled on the low-level interface of CaP-X: SAM~3 segmentation \citep{carion2026sam3} and Contact-GraspNet grasps \citep{sundermeyer2021contactgraspnet} through the services of CaP-X, depth, inverse kinematics, joint motions, whole-body motions planned around the observed scene with our planner, and the gripper.
Like \method{}, they get the starting images and notes on the conventions of the tasks, and like their originals, they get no function for articulated parts.

\emph{ReKep} \citep{huang2024rekep} marks keypoints on an image and has the VLM write the task as stages of constraints on them.
A solver turns these constraints into sub-goals and paths of the end effector and re-plans as the keypoints move.
As in its release, the simulator tracks the keypoints and the base stays in place.

We also compare with two VLAs trained on demonstrations of these tasks, the RoboCasa365 policies \citep{nasiriany2026robocasa365} of \emph{GR00T N1.5} \citep{bjorck2025gr00t}, which we run three times from the pretraining checkpoint, and of \emph{$\pi_{0.5}$} \citep{black2025pi05}, which we run from the multitask checkpoint.
We run them in our scenes and score them at the end of the episode, like the other methods.

\textbf{VLM and run-to-run variation.}
We run every zero-shot method with GPT-6.1-Sol at medium reasoning effort through the Codex CLI.
In \cref{sec:exp_sketch,sec:exp_failures}, we analyze two runs of \method{} with GPT-6-Astra, with which it succeeds as often (\cref{tab:vlm_final}).
Finally, we use simulated depth as calibrated metric depth.
Contact-rich episodes are chaotic, and VLM sampling adds to this variation.
Two runs of \method{} differ in 12 of the 90 episodes, so differences of two or three episodes in 90 are within run-to-run variation, and we therefore compare methods on paired episodes.

\subsection{Comparison with Zero-Shot Agents and VLAs}
\label{sec:exp_main}

\textbf{Kitchen tasks.}
As shown in \cref{tab:main}, \method{} succeeds in 82 of the 90 kitchen episodes with GPT-6.1-Sol (91\%), as often as with GPT-6-Astra.
It needs 5.4 VLM calls and 88k input tokens per episode, because the VLM answers a few questions about the world instead of iterating on actions, and it wrongly declared success in only 2 of its 81 claimed successes.
An episode takes 7.1 minutes on average (\cref{tab:time}).
The VLM thinks for 1.5 minutes and building the sketch takes another 0.6, mostly to move the camera for the views that the VLM requests, while planning and execution take 2.6 and 2.4 minutes.

In contrast, Code as Policies succeeds in only 8 episodes with the same VLM, and in none that \method{} fails (exact McNemar test on the paired episodes, $p < 10^{-21}$).
We find that most of its programs run to the end and miss, usually because they act on the wrong part or at the wrong point of it (\cref{sec:exp_failures}).
In its failed episodes, inverse kinematics finds no exact solution for 58\% of the targets, but we find that 89\% of these failures come from episodes whose program had already picked a wrong target.
The other 16 failures are program errors, most of them undefined names in helper functions that the model wrote.
Code as Policies reads few tokens (31k per episode) because it never sees an image, and its episodes take only 1.9 minutes because its program runs once, without feedback.

Feedback helps CaP-Agent0, which sees the output of its programs and a description by a VLM of how the scene changed.
With this feedback, it succeeds in 30 episodes, almost four times as often as Code as Policies ($p < 10^{-4}$), but in only one that \method{} fails ($p < 10^{-14}$).
It also picks wrong targets less often, since a wrong model of the task causes 14 of the 60 failures, compared with 55 of 82 for Code as Policies.
In its failed episodes, inverse kinematics finds an exact solution for 94\% of the targets, likely because the agent first places the base with \texttt{goto\_pose}, which it calls 13 times per failed episode and Code as Policies once.

Instead, CaP-Agent0 fails because it does not finish.
Of its 60 failures, 53 use all eleven programs, with 38 pointing requests per failed episode on average, and the other 7 end with a wrong claim of success.
We observe that it never solves a lever or the spout and solves doors only twice, which are the tasks where a program must compute the arc of the part.
In contrast, it succeeds in 12 of the 15 button episodes and in 7 of the 15 knob episodes, likely because a press or a turn at one point suffices there.
It makes 15 times as many VLM calls as \method{}, reads 10 times as many tokens and writes 25 times as many.
An episode takes 17 minutes, 2.4 times as long as with \method{}, and the VLM thinks for 13 of them, 5 of which go to pointing requests.
Concurrent work reports 61\% for a general-purpose coding agent with Claude Opus~5 on 18 atomic RoboCasa365 tasks, under its own task set and protocol \citep{chu2026openrua}.

ReKep succeeds in only 5 episodes, all of which \method{} also solves.
The released parser fails on 14 of its programs, so these episodes end before any motion (\cref{app:baselines}).
In 60 of the other 76 episodes, the solver backtracks to an earlier stage, and we traced most of these backtracks to grasps that did not hold, and 52 episodes end at the time limit after about 50 backtracks.
As a result, an episode of ReKep runs for at least 97 minutes on average, most of it in inverse kinematics.
Grasping is physical here, whereas the released code attaches a grasped object to the gripper.
However, most of its programs would fail even with a firm grasp, because their keypoints or constraints are wrong (\cref{sec:exp_failures}).

The two VLAs, which were trained on demonstrations of these tasks, succeed in 40 (GR00T, mean of three runs) and 35 ($\pi_{0.5}$) episodes, about half as often as \method{}.
They are also the fastest methods, at about one minute per episode, since they need no VLM calls and no planning.
GR00T does best on drawers, doors and levers, whose motions look alike across kitchens, and worst on buttons, knobs and pick-and-place, likely because there it must find a small part or a specific object.
It also often reaches a goal and then undoes it, and it would score 52--59 of 90 in its runs if success at any step counted.
$\pi_{0.5}$ fails almost every button and knob episode but places objects more often than GR00T (9 of 15, compared with 4.3).

\begin{table}[t]
  \caption{LIBERO-PRO, 150 episodes: position (pos.) and task perturbations of three suites, five tasks with five initial states each (successes of 25), with VLM calls and minutes per episode as in \cref{tab:main}. \emph{False}: episodes declared solved that were not. Every method that calls a VLM uses GPT-6.1-Sol, and the last row is \method{} with GPT-6-Astra. CaP-X runs as released, with its agent CaP-Agent0 and its single-turn tier with low-level functions (S3), which \cref{app:baselines} describes with our coding agent.}
  \label{tab:libero}
  \centering
  \small
  \setlength{\tabcolsep}{3pt}
  {\begin{tabular}{lcccccccccc}
    \toprule
    & \multicolumn{2}{c}{object} & \multicolumn{2}{c}{goal} & \multicolumn{2}{c}{spatial} & & & & \\
    method & pos. & task & pos. & task & pos. & task & all & calls & time & false \\
    \midrule
    $\pi_{0.5}$ \citep{black2025pi05} & 4 & 0 & 9 & 7 & 6 & 16 & 42 & -- & 0.3 & -- \\
    CaP-X, single turn \citep{fu2026capx} & 0 & 0 & 1 & 0 & 0 & 0 & 1 & 1.9 & 1.6 & -- \\
    CaP-Agent0 (CaP-X) \citep{fu2026capx} & 4 & 3 & 4 & 1 & 8 & 3 & 23 & 48 & 13.1 & 1 \\
    our coding agent & 25 & 25 & 25 & 23 & 24 & 25 & \textbf{147} & 5.8 & 0.9 & 1 \\
    \method{} & 25 & 25 & 23 & 24 & 24 & 25 & 146 & 3.2 & 1.4 & 1 \\
    \midrule
    \method{}, GPT-6-Astra & 25 & 25 & 24 & 24 & 24 & 25 & \textbf{147} & 3.3 & 2.4 & 0 \\
    \bottomrule
  \end{tabular}}
  \vspace{-3mm}
\end{table}

\textbf{Tabletop tasks.}
As shown in \cref{tab:libero}, the tabletop instance of \method{} succeeds in 146 of the 150 episodes of LIBERO-PRO (97\%), about as often as with GPT-6-Astra and as our coding agent, which both succeed in 147 (\cref{app:baselines}).
On a tabletop, depth clusters already separate the objects, so selecting among them suffices, unlike in a kitchen (\cref{sec:exp_ablation}).
\method{} needs 3.2 VLM calls and 1.4 minutes per episode.
Our coding agent needs 5.8 calls but only 0.9 minutes, because it does little besides calling the VLM, whereas \method{} updates its clusters at every keyframe and simulates the robot while the VLM thinks (\cref{app:runtime}).
Two of its four failures run out of calls or time while clearing the way for a drawer, and in a third the fitted model of a knob keeps absorbing the base of the knob.
In the fourth, the simulator ejects the plate, which also defeats the coding agent, and \method{} then puts the bowl on the stove and declares success.

We also run the released code of CaP-X (\cref{app:baselines}).
Its agent CaP-Agent0 succeeds in 23 episodes (15\%), within the 12--26\% per suite that CaP-X reports for it with other models \citep{fu2026capx}.
It succeeds in none of the episodes that \method{} fails, makes 15 times as many VLM calls and takes 13 minutes per episode.
The single-turn tier of CaP-X succeeds only once.

We find that both are misled by the interface of CaP-X.
Its documentation says that inverse kinematics solves for the hand and leaves the offset from the hand to the fingertips to the caller, although the inverse kinematics already applies this offset.
Every single-turn program and the first program of most CaP-Agent0 episodes therefore apply the offset again, so that the fingers close 10\,cm short of the grasp.
CaP-Agent0 sees the result and corrects the offset in 22 of its 23 successes.
However, about a third of the motion checks that it writes compare the commanded target with the reported pose of the hand, so they stop the program even after a correct motion.
All but one of its failed episodes run out of the 8,000 simulation steps that CaP-X allows or use all eleven programs.
ASPIRE reports 72\% after debugging the program of each task with the success check \citep{lu2026aspire}, and concurrent work reports 87\% for OpenRUA with up to 500 turns of Claude Opus~5 on four suites \citep{chu2026openrua}.

$\pi_{0.5}$, trained on LIBERO, succeeds in 42 episodes.
On the position perturbations, it succeeds in 16--36\% of the episodes per suite, close to the published 17--38\% \citep{zhou2025liberopro}.
Told the perturbed task, it also solves 23 of the 75 task-perturbed episodes, far more than the at most 1\% published for VLAs.
This difference likely comes from the instruction, which the LIBERO client of openpi derives from the name of the task, and LIBERO-PRO keeps the name of the original task.

\subsection{Sketch Accuracy and Ablations}
\label{sec:exp_sketch}
\label{sec:exp_ablation}

\begin{table}[t]
  \caption{Sketch accuracy (first sketch correct) and task success when one part of construction changes, as reference $\to$ variant in \%. Only the first block uses the final system. The others use earlier versions on other episodes (\cref{app:early_minimal,app:grounding,app:repeats,app:vlms}), so each row compares only with its own reference. $^\dagger$Recorded VLM answers replayed without execution. $^\ddagger$Right part bound, over the episodes that each arm finished.}
  \label{tab:ablation}
  \centering
  \small
  \setlength{\tabcolsep}{3.5pt}
  \begin{tabular}{lcc}
    \toprule
    change & correct & success \\
    \midrule
    \multicolumn{3}{l}{\emph{final system, 90 episodes}} \\
    no overlay check$^\dagger$ & 80 $\to$ 58 & -- \\
    \multicolumn{3}{l}{\emph{grounding, 50 ep., 2 runs}} \\
    masks instead of points & 71 $\to$ 46 & 77 $\to$ 65 \\
    hybrid: masks for panels & 71 $\to$ 75 & 77 $\to$ 75 \\
    \multicolumn{3}{l}{\emph{inventory, 50 ep., mask requests}} \\
    selection$^\ddagger$ & 45 $\to$ 63 & 65 $\to$ 41 \\
    \multicolumn{3}{l}{\emph{checks and hypotheses, 60 ep.}} \\
    no requested views & 69 $\to$ 71 & 78 $\to$ 80 \\
    single joint hypothesis & 69 $\to$ 66 & 78 $\to$ 78 \\
    no overlay check & 69 $\to$ 59 & 78 $\to$ 76 \\
    \multicolumn{3}{l}{\emph{VLM}} \\
    Claude Sonnet 5.5 (60 ep.) & 69 $\to$ 38 & 78 $\to$ 61 \\
    Gemini 4 Argon (30 ep.) & 77 $\to$ 60 & 63 $\to$ 57 \\
    \bottomrule
  \end{tabular}
  \vspace{-3mm}
\end{table}

\begin{table}[t]
  \caption{\method{} with different VLMs and reasoning efforts on the 90 kitchen episodes: successes, first sketches that are correct and that bind the right part, VLM calls per episode, and input, output and reasoning tokens per episode.}
  \label{tab:vlm_final}
  \centering
  \small
  \setlength{\tabcolsep}{3pt}
  {\begin{tabular}{lccccccc}
    \toprule
    VLM and effort & success & correct & part & calls & input & output & reasoning \\
    \midrule
    GPT-6-Astra medium, run 1 & 82 & 72 & 88 & 5.5 & 90k & 1.4k & 0.3k  \\
    GPT-6-Astra low & 77 & 69 & 84 & 5.3 & 86k & 1.2k & 0.2k \\
    GPT-6.1-Sol medium & 82 & 68 & 87 & 5.4 & 88k & 1.8k & 0.7k \\
    \bottomrule
  \end{tabular}}
  \vspace{-3mm}
\end{table}

We score the first sketch of each episode against the ground truth of the simulator (\cref{app:accuracy}).
A sketch is \emph{correct} when it binds the right part with the right joint type, its axis is within $10^\circ$, its interaction point within 2\,cm, the path of that point within 3\,cm of the true path, and its goal inside the success range of the task.

\textbf{The final sketches.}
In the two runs with GPT-6-Astra, the sketch binds the right part in 97\% of episodes and is fully correct in 81\% (\cref{tab:accuracy} gives each component).
The sketches are small.
The median sketch has one entity and two primitives and needs two requested views on average, in kitchens of over a thousand geometric bodies.
The VLM hedges where the mechanism is hidden.
It keeps several joint hypotheses for every knob and lever, for about a third of the doors and drawers, and for no button.
For the parts whose mechanism the images cannot show, the true joint is among the hypotheses for 82 of 89 joints and ranked first for 73 of them.

\textbf{What makes sketches accurate.}
In \cref{tab:ablation}, we change one part of construction at a time and find that grounding matters most.
As described in \cref{sec:construction}, a knob or a button is one snapped point and a lever two, whereas a mask misses small parts and does not show which way a part turns.
As a result, sketches grounded by masks are correct in 46\% of episodes, compared with 71\% for points.
Masks place the hinge edge of a large panel better, but a hybrid that uses them for panels does not succeed more often (\cref{app:grounding}).

Binding the right part is also not enough.
In selection-based construction, a backend segments the images automatically and fits a primitive to every mask, and the VLM picks among the numbered candidates.
This variant binds the right part at least as often as requests grounded by masks, yet it succeeds less often, and in 7 of the 20 failures no plan is feasible (\cref{app:early_minimal}).
Nor does modelling more help.
A sketch of every articulated part in view takes five times as long to build without succeeding more often, and a coarse object-level inventory fails almost always, since no plan can reach a small part that the inventory does not contain.

Among the checks, the overlay check matters most for accuracy.
When we replay the recorded VLM answers of the two runs and accept the first assembled sketch unchecked, correctness falls from 80\% to 58\%, and the share of sketches that bind the right part falls from 97\% to 86\% (\cref{app:accuracy}).
Requested views and extra hypotheses barely change correctness, and their effect on success is within run-to-run variation.
In our experiments, the VLM sets the ceiling.
Claude Sonnet~5.5 builds correct sketches in 38\% of episodes and Gemini~4 Argon in 60\%, compared with 69\% and 77\% for GPT-6 on the same episodes, and success falls with them (\cref{app:vlms}).
As shown in \cref{tab:vlm_final}, however, the choice between GPT-6-Astra and GPT-6.1-Sol matters little.
GPT-6.1-Sol succeeds as often (it gains 3 episodes and loses 3), although its first sketch is fully correct less often (68 compared with 72 of 90, $p = 0.34$).
Since it binds the right part just as often, its extra errors lie in the joint or the goal, and revisions and fresh sketches repair most of them: 17 of the 22 wrong first sketches still end in success.
At low reasoning effort, GPT-6-Astra loses 5 successes ($p = 0.18$) and saves few tokens.

\textbf{Accuracy and success.}
Success follows 93\% of correct first sketches and 77\% of wrong ones, because revisions and fresh sketches after failed attempts repair most wrong sketches.
Joint hypotheses rarely decide an episode in the final system, whose first hypothesis is usually right.
With an earlier version, whose first faucet hypothesis often had the wrong sense, the second hypothesis made 12 of 15 such faucet episodes succeed.
Sketches also need not be exact.
With an earlier executor, the same planner succeeded in 62 of 81 articulated episodes with ground-truth models and in 63 and 67 with sketches built by the VLM.
This is because closed-loop execution follows the real link and refits the joint as it goes, so the hand followed the true path within 0.3\,cm (median) even where the path of the sketch was 3--10\,cm off.
The role of the sketch is therefore to bind the right part and mechanism and to give a feasible way in.

\textbf{Execution components.}
After a failed attempt, a fresh sketch often succeeds.
When we replayed the recorded first attempts of a run without restarts and ran only the restarts live, success rose from 63 to 72 of 90 and no success was lost.
The final check removes false successes without losing true ones, and the verifier cut false successes from six episodes to one at unchanged success on a separate set of 30 episodes (\cref{app:repeats}).
Planning the contact together with the motion raised success on the 90 episodes from 55 to 62 with the outputs of the VLM replayed, and it solved 11 of 20 recorded contact failures, compared with 2 when the contact was chosen first.

\subsection{Failure Modes}
\label{sec:exp_failures}

\begin{table}[t]
  \caption{Root causes of the failures on the kitchen tasks, as shares of the failures of each method (\cref{app:failures}). \emph{World model}: the model of the task was wrong (for \method{} the sketch, for the code-writing agents the part, mechanism and goal that their programs assume, and for ReKep its keypoints and constraints). \emph{Execution}: the model was right but the motion failed. \emph{Other}: program and parser errors. \method{}: the two runs with GPT-6-Astra.}
  \label{tab:failmodes}
  \centering
  \small
  \setlength{\tabcolsep}{3pt}
  {\begin{tabular}{lcccc}
    \toprule
    & ours & CaP & CaP-Agent0 & ReKep \\
    \midrule
    failures (episodes) & 18/180 & 82/90 & 60/90 & 85/90 \\
    world model wrong & 11\% & 67\% & 23\% & 78\% \\
    \quad wrong part or contact site & 11\% & 52\% & 5\% & 56\% \\
    \quad wrong mechanism or direction & 0\% & 5\% & 13\% & 13\% \\
    \quad wrong goal or end state & 0\% & 10\% & 5\% & 8\% \\
    execution failed & 89\% & 13\% & 77\% & 6\% \\
    \quad no reachable contact & 17\% & 5\% & 23\% & 0\% \\
    \quad arm cannot follow the part & 0\% & 4\% & 13\% & 1\% \\
    \quad slip, small or wide part & 56\% & 5\% & 22\% & 2\% \\
    \quad placement off target & 11\% & 0\% & 0\% & 0\% \\
    \quad stopped early, estimate lost & 6\% & 0\% & 18\% & 2\% \\
    other & 0\% & 20\% & 0\% & 16\% \\
    \bottomrule
  \end{tabular}}
  \vspace{-3mm}
\end{table}

\begin{figure*}[t]
  \centering
  \includegraphics[width=\textwidth]{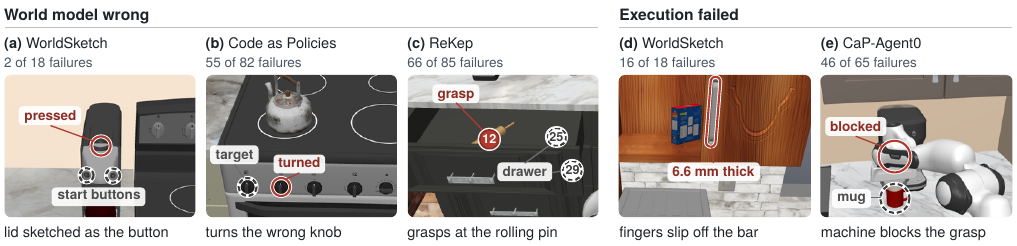}
  \caption{Failure cases on the kitchen tasks: for each method the most common cause in its main failure class (\cref{tab:failmodes}), and in (a) a sketch error of \method{}. Red marks show the error, and dashed grey marks the right part. (a) The VLM sketches the lid of the coffee machine as its start button. (b) Code as Policies turns the rear-left knob instead of the front-left one. (c) ReKep grasps at a keypoint on a rolling pin inside the drawer. (d) The fingers of \method{} slip off a cabinet bar 6.6\,mm thick (the frame is from after the slips). (e) The espresso head, hidden by the hand, blocks the grasps of CaP-Agent0 at the mug.}
  \label{fig:failures}
  \vspace{-3mm}
\end{figure*}

We sort the failures by root cause into three classes (\cref{tab:failmodes}): the model of the task was wrong, execution failed although the model was right, or something else went wrong.
\Cref{fig:failures} shows the most common cause of failure for each method.
For \method{}, only 2 of the 18 failures of the two runs come from the sketch.
In one, the VLM bound the lid of a coffee machine instead of its start button, and in the other it placed a mug at the drip tray, 10\,cm from the dispenser.
The other 16 failures happen in execution, mostly at the limits of the gripper (\cref{tab:failures}).
For example, the fingers slip off small parts or cannot enter them, such as thin door bars, a faucet lever or a recessed kettle button, and objects slip out of the hand or are wider than its 7.8\,cm opening.
Six of these 18 failures were declared successes, and 12 of the 15 failing episodes failed in only one of the two runs, which suggests that most failures are not systematic.

In contrast, Code as Policies and ReKep mostly act on a wrong model of the task (67\% and 78\% of their failures), most often on the wrong part or at the wrong point.
For example, Code as Policies resolves ``microwave door handle'' to the freezer handle of the fridge, 1.07\,m away, and ReKep binds its grasp to a keypoint on another part in 34 episodes, in 33 of which the image offered a keypoint on the right part.
As discussed in \cref{sec:exp_main}, CaP-Agent0 makes fewer such errors and fails mostly in execution (77\%).
Fixtures block its contact poses (14 episodes), grasps slip (13), its programs stop working motions with their own force checks (11), and the arm cannot follow a part while the base stays in place (8).
In 16 of these failures, its programs put the closing direction of the fingers on the wrong axis of the hand frame, although the interface documents it.

\subsection{Navigation}
\label{sec:exp_nav}

\begin{table}[t]
  \caption{Zero-shot navigation on R2R-CE val-unseen. Upper block: the 100 Open-Nav episodes, with published results. MIP ran GPT-6.1-Sol at low reasoning effort and reports only its SR. \method{} uses GPT-6.1-Sol at medium effort (mean of three runs, s.d.\ of SR 1.2). Lower block: 376 episodes on paths unused in development (one run). SR, SPL (success weighted by path length), OSR (oracle success) and nDTW (fidelity to the reference path) in \%, NE (final distance to the goal) in metres.}
  \label{tab:nav}
  \centering
  \small
  \setlength{\tabcolsep}{3pt}
  {\begin{tabular}{llccccc}
    \toprule
    method & VLM & SR & SPL & NE & OSR & nDTW \\
    \midrule
    Open-Nav \citep{qiao2025opennav} & GPT-4 & 19 & 16.1 & 6.70 & 23 & 45.8 \\
    SmartWay \citep{shi2025smartway} & GPT-4o & 29 & 22.5 & 7.01 & 51 & -- \\
    Three-Step Nav \citep{zheng2026threestepnav} & GPT-5 & 34 & 29.1 & 5.87 & 39 & 57.7 \\
    STRIDER \citep{he2025strider} & GPT-4o & 35 & 30.3 & 6.91 & 39 & 51.8 \\
    LaViRA \citep{ding2026lavira} & Gemini 2.5 Pro & 38.3 & 28.3 & 6.54 & 48.7 & -- \\
    EvoNav \citep{dai2026evonav} & Gemini 2.5 Pro & 43 & 37.8 & 5.04 & 51 & 62.4 \\
    AgenticNav \citep{li2026agenticnav} & Gemini 3.7 Flash & 76 & 66.5 & 3.48 & 78 & 69.4 \\
    MIP \citep{zhou2026mip} & GPT-6.1-Sol & 75.0 & -- & -- & -- & -- \\
    MIP \citep{zhou2026mip} & Fable 5.1 & 83.0 & 71.3 & 2.70 & 86.7 & -- \\
    \citet{dai2026astra} & GPT-6-Astra & 75.7 & 65.6 & 3.0 & 80.7 & 70.5 \\
    \method{} & GPT-6.1-Sol & 75.7 & 57.5 & 2.86 & 83.0 & 67.4 \\
    \midrule
    \method{} & GPT-6.1-Sol & 76.9 & 56.6 & 2.60 & 82.2 & 67.2 \\
    \bottomrule
  \end{tabular}}
  \vspace{-3mm}
\end{table}

We evaluate the navigation instance on R2R-CE \citep{anderson2018r2r,krantz2020vlnce} in Habitat \citep{savva2019habitat}, in the standard setting with discrete steps and turns, one forward RGB-D camera and no GPS or compass.
The robot estimates its pose by dead reckoning, which it corrects by registering depth with earlier scans.
As in the kitchen comparison, the VLM is GPT-6.1-Sol at medium effort (\cref{app:nav} gives the remaining details).
We test on 376 val-unseen episodes on paths that we never used in development and, with three runs, on the 100 episodes of \citet{qiao2025opennav}, on which most zero-shot methods report their results.

\method{} reaches the goal in 77\% of the unused episodes and in 76\% of the Open-Nav episodes (\cref{tab:nav}), or in 75\% without the 13 episodes that were in our development set.
This is far above methods built on earlier VLMs (19--43\%) and the same as AgenticNav and as agents that ask GPT-6 models for each action \citep{zhou2026mip,dai2026astra}.
In that interface, Claude models and GPT-6-Astra at its highest effort reach 81--83\%.
Because the planner takes the actions, \method{} calls the VLM 5.7 times per episode, whereas the step-by-step agent with the same VLM makes 30 calls and reads six times as many input tokens \citep{zhou2026mip}.
However, the trajectories of \method{} are longer (SPL 57\%, compared with 66\% for GPT-6-Astra).
Successful episodes with two or three VLM turns reach an SPL of 90--96\%, close to the 89\% of the planner when it is given the goal (\cref{app:nav}).
With eight or more turns, the SPL falls to 51\%, as the robot moves from one door or passage to the next and its path weaves along the route or detours into a wrong room.

We find that most failures come from the interpretation of the instruction and few from localization, since the pose error exceeds 0.5\,m in only 5 of 300 runs.
Of the 25 failures in the first run, which we classified by hand, 14 end at another room, door or landmark than the one meant, for example a second fire extinguisher further down the corridor, and 8 stop just outside the 3\,m radius.
Routes over stairs succeed as often as routes on one floor (81\% and 76\% of the unused episodes), likely because a staircase primitive lets the planner join the floors.

\section{Limitations and Conclusion}
\label{sec:conclusion}

\textbf{Limitations.}
We evaluated all methods in simulation, with simulated depth and calibration.
Real sensors will add noise to every measurement that the sketch is built from, and in replays the system tolerated 0.5\,cm of depth noise but not 1\,cm (\cref{app:robust}).
 Many contact strategies of the executor were designed for the appliances of RoboCasa, such as a fast tug that frees a drop-down door from its detent or pushing a wire rack in by its front rail, so new mechanisms will likely need new contact strategies.
In addition, our sketches cover rigid parts with one prismatic or revolute joint, and deformable objects, chains of joints and dynamic tasks are out of scope.
The VLM is a proprietary model whose calls take 15\,s (median), so the five calls of an episode add about 1.5 minutes to its 7 minutes.
Most remaining failures are grips on small or thin parts and placements that need centimetre accuracy, which better contact primitives must address.

\textbf{Conclusion.}
\method{} asks a VLM for a sketch of the world instead of an action.
The VLM decides what matters and what the mechanism is, measurements supply the geometry, and a planner computes the motion.
A minimal kinematic sketch built in this way is small, can be checked against the images and can be tested by execution.
In our experiments, it outperformed two trained VLAs and three zero-shot methods that ask the VLM for programs or constraints on RoboCasa kitchen tasks, solved 97\% of the perturbed tabletop tasks of LIBERO-PRO, and in navigation reached as many goals as step-by-step agents with a fifth of their VLM calls.
We believe that the open problems now lie in contact-rich execution and in mechanisms that no image reveals, where the robot must find out by trying.

\bibliography{refs}
\bibliographystyle{plainnat}

\newpage
\appendix




\section{Tasks}
\label{app:bench}

\begin{table}[h]
  \caption{The RoboCasa kitchen tasks: five atomic tasks per family, three episodes each (seeds 1000--1089, RoboCasa365's default kitchens). Tasks in italics were drawn with a fixed random seed from the RoboCasa365 atomic tasks that act on the same kind of part as their family.}
  \label{tab:bench_rc}
  \centering
  \small
  \begin{tabular}{@{}lp{13.4cm}@{}}
    \toprule
    family & tasks \\
    \midrule
    drawers, racks & OpenDrawer, CloseDrawer, \emph{SlideOvenRack}, \emph{SlideToasterOvenRack}, \emph{CloseFridgeDrawer} \\
    doors          & OpenCabinet, CloseCabinet, OpenMicrowave, CloseMicrowave, OpenToasterOvenDoor \\
    knobs          & TurnOnStove, TurnOffStove, \emph{PreheatOven}, \emph{TurnOnToasterOven}, \emph{LowerHeat} \\
    buttons        & TurnOnMicrowave, TurnOffMicrowave, \emph{StartCoffeeMachine}, \emph{TurnOnBlender}, \emph{OpenElectricKettleLid} \\
    levers, spout  & TurnOnSinkFaucet, TurnOffSinkFaucet, TurnSinkSpout, \emph{TurnOnElectricKettle}, \emph{AdjustWaterTemperature} \\
    pick \& place  & PickPlaceCounterToSink, \emph{PickPlaceMicrowaveToCounter}, \emph{PickPlaceCabinetToCounter}, \emph{PickPlaceCounterToDrawer}, \emph{CoffeeSetupMug} \\
    \bottomrule
  \end{tabular}
\end{table}

Each family starts from common RoboCasa365 tasks on its kind of part (upright roman in \cref{tab:bench_rc}) and is filled to five with tasks drawn at random from a pool: for drawers and racks, the fridge drawers and the dishwasher, oven and toaster-oven racks; for knobs, the other stove, oven and toaster-oven knob tasks; for buttons, every other task whose instruction says to press a button; for levers, every other task whose instruction names a lever or the water temperature; for pick and place, ten pick-and-place tasks between counters, appliances, cabinets and drawers.
The door family has five common tasks.
The checks of RoboCasa also require the gripper to end away from a pressed button (15\,cm for the microwave and coffee-machine buttons) and from a placed object (25\,cm), which the instruction does not state.
For the zero-shot methods, we count a press of these buttons once the appliance has switched (\cref{sec:exp_setup}).
Placements keep the full check, because there the withdrawal shows that the object was released.

The LIBERO-PRO tasks use the suites \texttt{libero\_\{object, goal, spatial\}\_\{swap, task\}} (position and task perturbations).
Five tasks per original suite were drawn with a fixed random seed and used for both of its perturbations: object 0, 2, 4, 6, 9; goal 2, 3, 4, 6, 7; spatial 1, 2, 3, 7, 9 (task indices of the released suites).
Each task is run from initial states 10--14 of the released files.
Success is the goal predicate of LIBERO in the final state, with no step limit. \method{} may use 12 VLM calls and 300\,s of simulated time per episode, with the VLM's latency charged to the robot's clock, and the coding agent 20 turns and 600\,s of motion.

\section{Sketch Construction: Tools, Sketch Language and Examples}
\label{app:tools}

\textbf{One turn.}
The first message holds a fixed instruction text, the task instruction and four images: the three head cameras and an upward wrist view of the wall cabinets, each resized to $768 \times 576$ with pixel ticks on its borders.
Every turn the VLM answers in one structured format with five fields: its reasoning, point requests, view requests, the sketch (or nothing, to keep the last one) and a decision (continue, accept or fail; a restarted session may also answer that the task is done).
The backend lifts the points, takes the views, rebuilds the sketch from the latest points, checks every primitive against depth and draws the sketch into the images.
Its reply lists each point's 3-D position and fit, the new views, the sketch as built with its notes and errors, the depth check per primitive and image (agrees, sees through, hidden), and up to four overlays.
All turns of a sketch run in one resumed session, so a turn sends only what is new.
Acceptance is the decision of the VLM, and no threshold accepts a sketch.

\begin{table}[h]
  \caption{The tools the VLM calls while sketching (the request fields of its answer) and how the backend grounds them. \emph{Uses}: requests over the 90 episodes of one run.}
  \label{tab:tools}
  \centering
  \small
  \setlength{\tabcolsep}{4pt}
  \begin{tabular}{@{}p{2.0cm}p{3.1cm}p{6.6cm}p{2.9cm}r@{}}
    \toprule
    tool & the VLM writes & grounding & returned & uses \\
    \midrule
    surface point & name, image, pixel & a plane fitted to the depth around the pixel (6\,mm RANSAC); the point is where the pixel's ray meets it & 3-D point, normal & 78 \\
    part point & name, image, pixel & a plane fitted to the ring around the pixel; depth more than 6\,mm in front of it is clustered, and the nearest cluster's centroid is the point & point, protrusion, size, long axis & 195 \\
    corner & name, image, pixel & depth along the pixel's ray; when a panel's corners come from one image, they are placed on the plane fitted inside them, preferring a front plane (glass) & point & 291 \\
    raw point & name, image, pixel & median depth along the pixel's ray & point, normal & 199 \\
    wrist view & pixel, distance (0.4, 0.65 or 0.9\,m) & the wrist camera looks at the 3-D point under the pixel, from a pose with a free line of sight that the arm can reach & new image & 83 \\
    label view & pixel & the wrist camera straight in front of the surface under the pixel, about 45\,cm away (the base backs off if needed) & high-resolution image & 35 \\
    crop & image, box & a zoom into the full-resolution frame & new image & 42 \\
    sketch & the sketch in the format below & built from the named points, depth-checked and drawn into every image & the sketch as built, notes, depth check, overlays & 253 \\
    \bottomrule
  \end{tabular}
\end{table}

\textbf{Sketch language.}
A sketch lists parts, each with a link (a panel given by three or four corners, a bar given by two ends, or a box), a part where the robot acts (a cylinder given by two ends or by a centre, an axis and a length; a disk; a button; a box; or none, to push the link), joint hypotheses with priors, and optionally the current state and an effect region.
A joint is revolute or prismatic, and its axis is given by two points, by an edge of the link's panel (\texttt{edge:left}, \texttt{right}, \texttt{top}, \texttt{bottom}, \texttt{near}, \texttt{far}), or by a direction word (\texttt{vertical}, \texttt{horizontal}, \texttt{normal}, \texttt{left}, \texttt{toward}, \texttt{up}) through a point.
A goal names a motion (open, close, clockwise, counter-clockwise, press, or a direction in which the part moves), a rough amount, and whether to continue until the stop. It may instead name where the part should end.
The state of a part to be closed is given by three points on the surface it closes onto, or by how far it is open.
Pick-and-place tasks add objects: the object to move (a box or a sphere at a point) and the surface to place it on.
The VLM writes names, not numbers: the backend computes axes, origins, sizes and the sense of every goal from the measured points.
It also adds hypotheses the VLM did not give (\cref{app:details}).

\textbf{Example: the toaster oven door of \cref{fig:overview}} (excerpts, verbatim; positions in metres in the robot's frame).
\begin{itemize}
  \setlength\itemsep{1pt}
  \item \emph{VLM, turn 1}: ``The toaster oven has a closed, bottom-hinged front door with a horizontal handle along its top edge.'' Points in the centre head image: four corners of the door and two part points at the ends of the handle. Sketch: the door as a panel on the four corners, the handle as a cylinder between its ends, one hypothesis (revolute about \texttt{edge:bottom}, open $90^\circ$ until the stop), state open $0^\circ$.
  \item \emph{Backend}: the corners lie at $x \approx 0.80$\,m; \texttt{handle\_left} ``protrudes 2.1\,cm, size $5.8 \times 3.2$\,cm, continues beyond the 20\,px search''. ``panel door\_panel: only 35\% of the depth inside its corners lies on one plane''. Depth check of the panel in the centre image: 33\% agrees, 7\% sees through, 60\% hidden; the handle agrees everywhere. Four overlays; in the centre one the panel lies flat behind the oven and its goal pose stands up above it.
  \item \emph{VLM, turn 2}: ``The handle and bottom hinge align correctly, but the glass caused an incorrect panel-plane fit. Replace that panel with a hinge-to-handle bar while retaining the measured bottom hinge axis and downward opening motion.'' Sketch: the door as a bar from the bottom-left corner to the handle's left end, the hinge axis through the two bottom corners.
  \item \emph{Backend}: axis $[-0.008, -1.0, 0.023]$ through $[0.805, 0.089, 0.962]$; bar 24.9\,cm, handle 32.2\,cm; five hypotheses (the VLM's at 0.77 and four other axes at 0.06); the bar agrees with depth in 84\% of its pixels, the handle in all.
  \item \emph{VLM, turn 3}: accept: ``The simplified door link and handle align with the oven; the axis follows its bottom edge, and the goal arrow shows the handle moving outward and down to open it.''
\end{itemize}
Against ground truth, the accepted axis is off by $1.4^\circ$ and the hinge line by 3\,mm.
Execution tried the first hypothesis four times: a grasp opened the door $59^\circ$ and slipped; a grasp missed; a push reached the stop but the verifier, seeing no change in appearance, refuted it (belief $0.74 \rightarrow 0.22$); after the same push, the uncertain verdict went to the VLM, which judged the before-and-after image ``ACHIEVED The toaster oven door is lowered to its fully open, approximately horizontal position'' (belief $0.88$).
\Cref{fig:construction} shows two more constructions: a knob chosen by reading its icon in a label view, and a lever re-pointed in a close view and sketched with four joint hypotheses.

\section{Implementation Details}
\label{app:details}

\textbf{Construction.}
The VLM receives the instruction, the three head-camera images and an upward wrist view of the wall cabinets.
A sketch takes up to five turns in one resumed session, and an episode may use up to 14 VLM calls.
A panel whose corners do not enclose one plane is flagged in the reply of the backend.

\textbf{Outcome likelihoods.}
\Cref{tab:lik} lists $P(o \mid \rho)$ for the belief update.
``No contact'' carries no evidence.
The system adds the opposite sense of a guessed direction with prior $0.25$ relative to its hypothesis, each unused axis of a lever in both senses with $0.15/2$ of the top prior, and a press with $0.2$ of the top prior for a part under 6\,cm with no prismatic hypothesis and no effect region.

\begin{table}[h]
  \caption{Outcome likelihoods $P(o \mid \rho)$ used in the belief update.}
  \label{tab:lik}
  \centering
  \small
  \begin{tabular}{lccc}
    \toprule
    outcome $o$ & same & flipped & other \\
    \midrule
    verified                      & 0.80 & 0.01 & 0.03 \\
    satisfied, unconfirmed        & 0.10 & 0.05 & 0.10 \\
    moved without effect          & 0.04 & 0.25 & 0.40 \\
    moved the full amount, no effect & 0.15 & 0.25 & 0.40 \\
    moved along another joint     & 0.05 & 0.20 & 0.50 \\
    resisted                      & 0.04 & 0.60 & 0.40 \\
    moved partly                  & 0.06 & 0.09 & 0.07 \\
    no contact                    & 1.00 & 1.00 & 1.00 \\
    \bottomrule
  \end{tabular}
\end{table}

\textbf{Verifier.}
Snapshots are taken with the arm tucked, from the base pose of the before-snapshot (within 5\,mm and $0.3^\circ$).
An after-pixel counts as changed when its depth disagrees along the reprojected ray or its colour differs from the best match in a $5 \times 5$ neighbourhood, and isolated pixels are dropped.
The effect check keeps changed pixels whose 3-D points lie at most 12\,cm above the effect surface, in connected blobs. For water and flames, it also compares the amount of water- or flame-coloured pixels over the effect region before and after.
Missing force or missing effect never refutes a goal: a drawer slides at 0.2\,N, and a flame may be hidden under a pan.

\textbf{Planner.}
\label{app:planner}
\begin{itemize}
  \setlength\itemsep{0pt}
  \item \emph{Contacts.} Bar grasps: approach tilts $0, \pm20, \pm40, \pm65, \pm90^\circ$ about the bar, offsets $0$ and $\pm L/4$ on bars longer than 8\,cm, both finger directions, grip depth from the measured standoff. Knob grasps: six closing directions $30^\circ$ apart, tilts $0, \pm20, \pm35^\circ$, a nominal and a shallow depth. Button presses with the closed fingertips, tilted by $20$--$50^\circ$ when no straight press plans. Pushes on a grid of three to five points per face direction (about one per 15\,cm), at least 6\,cm from the part, where the motion points into the face (within $53^\circ$ of its inward normal). Edge pinches at five places per exposed edge, only after an earlier round on the link or when nothing else plans. The pre-contact pose, 12\,cm back along the approach (8--10\,cm for pushes), must lie in space a camera saw free. A pushing hand on a revolute link turns by half the joint angle.
  \item \emph{Sampling.} Joint samples every $6^\circ$ ($10^\circ$ for parts within 8\,cm of the axis) or 4\,cm. Placements: forward $-0.6$ to $0.6$\,m and sideways $-0.7$ to $0.7$\,m in 0.1\,m steps, yaw $0, \pm20, \pm40, \pm65, \pm90^\circ$, four torso heights (7{,}020 in all); those in reach with a free footprint are ranked, the best 240 are screened by inverse kinematics along five path samples, and the best 24 per contact cluster (contacts within 8\,cm) are paired with every contact of the cluster, with up to three arm branches per pair. A sample fails on inverse kinematics, an unnatural posture, a joint jump above 0.6\,rad, an arm sphere within 2\,cm of a map point or of the moving link, or (every third sample) the hand boxes meeting the map. Pairs are ranked by the fraction of the path followed, then by base motion, reach comfort, approach direction and grasp alignment; VAMP must find the approach. The budget is 8 work-seconds, counted in inverse-kinematics iterations and collision checks, so planning is deterministic.
  \item \emph{Collisions.} Collisions are checked between sphere and box models of the robot and the point cloud of $\C$ and of the moving link.
  \item \emph{Optimization.} The augmented-Lagrangian method takes Levenberg--Marquardt steps from seeds given by grasp tilts and by base placements from the sampling, and a solution becomes the plan only if its base pose and approach are feasible. The optimization runs first when a door is opened by its handle, and drawers and levers use it when sampling finds no plan. Residuals are scaled so that one unit is 1\,cm or 0.1\,rad. Bounds: base $\pm0.8$\,m and $\pm90^\circ$, tilt $\pm75^\circ$, joints 0.1\,rad inside their limits. Margins: 1\,cm for the arm spheres; 3.5\,cm from the map and 3\,cm from the swept link for the base; 4\,mm from the link, bar and posts for the hand, which may touch the link's face lightly at the end of the guarded approach (2.5\,mm) and while the fingers close (1\,mm); 6\,mm from the scene; at most 0.35\,rad per joint between samples. The hand is the palm, finger and wrist boxes of the robot's collision meshes. The bar is re-measured from depth as a rounded box held on its straight segment between its posts; the panel is a 2\,cm solid box plus its measured surface. Solver: at most 6 outer iterations (penalty $\times4$ each) of at most 20 Levenberg--Marquardt steps; a solution is accepted below 3\,mm (0.03\,rad) equality error and 1\,mm violation. Seeds: tilts $0, \pm30^\circ$ ($\pm60, \pm90^\circ$ added for levers) times two base placements per finger direction; the search stops after two valid solutions, checked by inverse kinematics halfway between samples. A prefix ends just before the first sample that the least-violating solution violates, or at 40\% of the motion. Sliding links also try a hook (closed fingers behind the bar), and a door whose grip gave way after it opened is hooked at its free edge. In one run of the kitchen tasks the optimizer planned 38 of 492 executed contacts (27 of 153 on doors); sampling planned the rest.
  \item \emph{Execution.} Increments of about 2\,cm along an arc ($5$--$20^\circ$), 3\,cm along a slide, $10^\circ$ for knobs. A step that advances less than 35\% of its increment stiffens the controller (position gain $150 \rightarrow 450$); two more end the round. The grip counts as lost when the finger gap falls under 4\,mm or changes by 1\,cm. A stop must lie within 3\,cm or $8^\circ$ of the re-observed link; stalls within the first 5\,cm or $12^\circ$, stalls where the grip gave way, and closing stalls well short of the closed state need a second contact. Force guards: 15\,N into contact, 25\,N on planned free motions. At most four rounds per sense; a contact that did not move the link, or whose way in was blocked, is not tried again.
  \item \emph{Objects.} An object's footprint is measured in depth as the points connected to the sketch's point above the surface it stands on, and their smallest gravity-aligned rectangle. It is grasped from above across the footprint's narrow side (or at a narrower section when it is too wide), or from the side under a shelf or in a microwave. The release is sought on the destination's surface near its centre, with the hand turned in $30^\circ$ steps, where the hand and the carried object have a free straight way in.
\end{itemize}

\textbf{Restarts.}
A restart drives the base back to the first observation pose, observes again (an upward wrist view and three head views), and starts a new VLM session with the construction prompt and a summary of the failed attempt: its sketch with all hypotheses, the last evidence lines and the reason it failed.
The new session may answer that the task is done.
The final check runs when the attempt ended with every goal satisfied but some of them rest on weak evidence. Its session is told which results the checks of the robot could not see.

\section{Runtime}
\label{app:runtime}

We time an episode from the start of the agent to the final success check, without building the scene, and we count each VLM call by the response time of the model.
We do not count the time that a request spent in our relay, which carried the requests of all methods to the Codex CLI on another machine and added 12--26\,s per call, depending on the load.
CaP-Agent0 queries its three candidate programs in parallel, so each turn costs its slowest candidate.
\Cref{tab:time} splits the time into stages.
We assign each stretch of the log of an episode to the stage of the event that ends it, and the simulated motion of the robot to execution, except for the motion that \method{} makes to observe the scene for its sketch, which counts towards building the model.

\begin{table}[h]
  \caption{Time per episode by stage, in seconds: VLM response time (\emph{think}), building the world model apart from the VLM (\emph{model}), motion planning and inverse kinematics (\emph{plan}), robot motion and outcome checks in the simulator (\emph{execute}), and the rest. Where the logs do not separate model, plan and execute, one value spans the three columns. The VLAs log only the total.}
  \label{tab:time}
  \centering
  \small
  \begin{tabular}{lrrrrrr}
    \toprule
    method & think & model & plan & execute & other & total \\
    \midrule
    \multicolumn{7}{l}{\emph{kitchen tasks}} \\
    \method{} & 91 & 37 & 157 & 143 & 0 & 429 \\
    \method{}, GPT-6-Astra & 78 & 36 & 112 & 149 & 0 & 374 \\
    Code as Policies & 56 & 25 & 3 & 33 & 0 & 117 \\
    CaP-Agent0 & 792 & 65 & 8 & 154 & 0 & 1019 \\
    ReKep & 35 & 30 & 5442 & 334 & 0 & $\geq$5842 \\
    $\pi_{0.5}$ & -- & -- & -- & -- & -- & 69 \\
    GR00T N1.5 & -- & -- & -- & -- & -- & 45 \\
    \midrule
    \multicolumn{7}{l}{\emph{LIBERO-PRO}} \\
    \method{} & 34 & 25 & 7 & 14 & 5 & 86 \\
    \method{}, GPT-6-Astra & 39 & 55 & 10 & 32 & 9 & 145 \\
    our coding agent & 44 & \multicolumn{3}{c}{8} & 1 & 53 \\
    CaP-Agent0 & 531 & \multicolumn{3}{c}{240} & 12 & 784 \\
    CaP-X, single turn & 40 & \multicolumn{3}{c}{46} & 11 & 97 \\
    $\pi_{0.5}$ & -- & -- & -- & -- & -- & 20 \\
    \bottomrule
  \end{tabular}
\end{table}

In the kitchen, an episode of \method{} with GPT-6.1-Sol takes 7.1 minutes on average (median 3.5) and 92\,s of simulated robot motion (median).
A VLM call takes 15\,s (median, and 7--31\,s for 80\% of the calls).
Building the sketch takes 37\,s, of which the robot spends about 28\,s moving to take the observations and views that the VLM requests.
Planning takes the largest share, mostly in the contact--motion optimization (135\,s per episode on average), which runs for every door opened by its handle and for the drawers and levers that sampling cannot plan.
Code as Policies and CaP-Agent0 spend 48\% and 78\% of their time in VLM calls, and CaP-Agent0 spends 5.4 minutes per episode on pointing requests alone.
ReKep spends 93\% of its time in its solvers, most of it in inverse kinematics.
The VLAs take 0.8 (GR00T) and 1.2 ($\pi_{0.5}$) minutes per episode, with policy servers shared by three or four clients, and their logs do not separate inference from simulation.
On LIBERO-PRO, the harness of \method{} simulates the robot while the VLM thinks, because it charges the latency of the VLM to the robot's clock, and it updates the clusters of the scene at every keyframe.
The run with GPT-6-Astra waited for a VLM lock that it shared with another run, and these waits were simulated as well, which adds about 30\,s per episode.

\section{Sketch Accuracy}
\label{app:accuracy}

The ground truth is read from RoboCasa's own success check of each episode: the parts and joints it tests, the joint ranges it accepts, any part it accepts instead (for example, a second start button), and the destination regions of pick-and-place tasks.

\begin{table}[h]
  \caption{Accuracy of the first sketch against ground truth in the two runs with GPT-6-Astra (\cref{tab:vlm_final}). \emph{Correct}: right part, joint type, axis ($\le 10^\circ$), interaction point ($\le 2$\,cm), path ($\le 3$\,cm) and goal inside the success range. Per-joint rates are over target joints; \emph{truth among hypotheses} is over joints of faucets, spouts, knobs and buttons.}
  \label{tab:accuracy}
  \centering
  \small
  \setlength{\tabcolsep}{4pt}
  \begin{tabular}{lrcccccc c}
    \toprule
    set & episodes & correct & type & axis & point & path & goal & truth among hypotheses \\
    \midrule
    kitchen tasks (2 runs) & 180 & 81\% & 97\% & 94\% & 88\% & 93\% & 92\% & 82/89 \\
    \bottomrule
  \end{tabular}
\end{table}

\begin{table}[h]
  \caption{Accuracy of the first model for three ways of grounding it, with an earlier version (420 episodes). Axis, pivot, path and direction are scored per joint.}
  \label{tab:accuracy_variants}
  \centering
  \small
  \setlength{\tabcolsep}{3pt}
  \begin{tabular}{lrcccccc}
    \toprule
    grounding & $n$ & correct & axis $\le 10^\circ$ & pivot (cm) & path $\le 3$\,cm & direction & success \\
    & & & & median / p90 & & right & given correct / not \\
    \midrule
    points    & 220 & 68\% & 92\% & 1.7 / 3.2  & 76\% & 92\% & 85\% / 66\% \\
    hybrid    & 100 & 75\% & 93\% & 1.5 / 3.1  & 80\% & 92\% & 77\% / 68\% \\
    masks     & 100 & 46\% & 93\% & 2.9 / 35.2 & 67\% & 73\% & 80\% / 52\% \\
    \bottomrule
  \end{tabular}
\end{table}

Mask-grounded models more often have the wrong direction of motion (27\% of cases) and a heavy tail of pivot errors, which segments do not constrain.

In the two runs of \cref{tab:accuracy}, success follows 135 of the 145 correct first sketches and 27 of the 35 wrong ones, and 3 of the 146 verified parts were verified under a hypothesis other than the first.
With the recorded VLM answers replayed, accepting the first assembled sketch without the overlay check loses 44 correct sketches and gains 5 on 173 episodes. An earlier version lost 10 points of correctness in the same way (\cref{tab:ablate}).

\section{Failure Causes}
\label{app:failures}

\begin{table}[h]
  \caption{Root causes of the 18 failures in the two runs with GPT-6-Astra (\cref{tab:vlm_final}), one cause per failure, from the logs and replays.}
  \label{tab:failures}
  \centering
  \small
  \begin{tabular}{p{4.2cm}cp{9.3cm}}
    \toprule
    cause & failures & examples \\
    \midrule
    small part the fingers slip off or cannot enter & 6 & 6.6\,mm cabinet bars inside a thicker visual handle; a small faucet lever; a toaster tray's thin lip; a kettle button recessed 1--2\,mm in the handle \\
    no reachable contact & 3 & a cabinet door open flat against the fridge; a plate deep inside a microwave \\
    wrong functional site in the sketch & 2 & the lid of a coffee machine pressed instead of its start button; a mug placed at the drip tray, 10\,cm from the dispenser \\
    placement settles off target & 2 & a mug released within 0.5\,cm of the dispenser settled 4.2--4.4\,cm away (limit 4\,cm) \\
    held object slipped out & 2 & a cup dropped during the carry; a mug lost at the release \\
    object wider than the gripper & 2 & a translucent container, 9.7\,cm wide, whose depth hits the shelf behind it \\
    joint estimate reset & 1 & a spout turned the wrong way, refuted, and its angle reset to the start \\
    \bottomrule
  \end{tabular}
\end{table}

Six of the 18 failures were declared successes: the coffee machine pressed at the wrong site, verified by force alone; the recessed kettle button; the mug at the drip tray, which ended with an empty revised sketch; the two mugs that settled just outside the limit, which the final check accepted; and the reset spout.

\textbf{Baselines.}
We classified every failed episode of Code as Policies, CaP-Agent0 and ReKep in the same way and checked each by hand.
We compare the part, mechanism and goal that a program or a constraint set assumes with the simulator's ground truth, read the outcome of each motion from the logs, and count the model when both the model and the motion were wrong.
A missing withdrawal from a placed object counts as a wrong goal only when the placement itself succeeded (one ReKep episode).
We count ReKep's 14 parser failures as other, although 11 of the 13 programs that the released parser cuts short would themselves have been wrong.

\section{Per-Task Results}
\label{app:pertask}

\begin{table}[h]
  \caption{Successes per task on the 90 kitchen episodes (of 3): \method{} with GPT-6-Astra (the run of \cref{tab:main}) and with GPT-6.1-Sol, Code as Policies (CaP), CaP-Agent0 (A0) and ReKep with GPT-6.1-Sol, $\pi_{0.5}$, and the multi-turn coding agent of \cref{app:baselines}.}
  \label{tab:pertask}
  \centering
  \scriptsize
  \setlength{\tabcolsep}{2.5pt}
  \resizebox{\textwidth}{!}{\begin{tabular}{lccccccc@{\hskip 12pt}lccccccc}
    \toprule
    & \multicolumn{2}{c}{ours} & & & & & & & \multicolumn{2}{c}{ours} & & & & & \\
    task & Astra & Sol & CaP & A0 & ReKep & $\pi_{0.5}$ & coding & task & Astra & Sol & CaP & A0 & ReKep & $\pi_{0.5}$ & coding \\
    \midrule
    OpenDrawer           & 3 & 3 & 1 & 2 & 0 & 2 & 3 & TurnOnMicrowave             & 3 & 3 & 0 & 3 & 0 & 1 & 3 \\
    CloseDrawer          & 3 & 3 & 1 & 2 & 0 & 2 & 3 & TurnOffMicrowave            & 3 & 3 & 1 & 3 & 0 & 0 & 3 \\
    SlideOvenRack        & 3 & 2 & 0 & 1 & 1 & 1 & 2 & StartCoffeeMachine          & 2 & 3 & 1 & 3 & 0 & 0 & 3 \\
    SlideToasterOvenRack & 3 & 3 & 0 & 0 & 0 & 2 & 1 & TurnOnBlender               & 3 & 3 & 0 & 1 & 0 & 0 & 0 \\
    CloseFridgeDrawer    & 3 & 3 & 1 & 1 & 1 & 3 & 2 & OpenElectricKettleLid       & 2 & 3 & 0 & 2 & 0 & 0 & 3 \\
    OpenCabinet          & 3 & 3 & 0 & 0 & 0 & 0 & 1 & TurnOnSinkFaucet            & 3 & 3 & 0 & 0 & 0 & 2 & 0 \\
    CloseCabinet         & 2 & 2 & 0 & 0 & 0 & 1 & 1 & TurnOffSinkFaucet           & 3 & 3 & 0 & 0 & 0 & 3 & 1 \\
    OpenMicrowave        & 3 & 3 & 0 & 1 & 0 & 1 & 3 & TurnSinkSpout               & 3 & 3 & 0 & 0 & 0 & 3 & 3 \\
    CloseMicrowave       & 3 & 3 & 0 & 0 & 0 & 3 & 3 & TurnOnElectricKettle        & 3 & 3 & 0 & 0 & 0 & 0 & 3 \\
    OpenToasterOvenDoor  & 3 & 3 & 0 & 1 & 0 & 0 & 1 & AdjustWaterTemperature      & 2 & 2 & 1 & 0 & 1 & 0 & 1 \\
    TurnOnStove          & 3 & 3 & 1 & 1 & 0 & 1 & 3 & PickPlaceCounterToSink      & 3 & 3 & 0 & 2 & 1 & 2 & 3 \\
    TurnOffStove         & 3 & 3 & 0 & 2 & 0 & 0 & 3 & PickPlaceMicrowaveToCounter & 2 & 2 & 0 & 0 & 0 & 2 & 1 \\
    PreheatOven          & 3 & 2 & 0 & 2 & 0 & 0 & 2 & PickPlaceCabinetToCounter   & 2 & 2 & 0 & 0 & 0 & 1 & 2 \\
    TurnOnToasterOven    & 3 & 3 & 1 & 2 & 1 & 1 & 3 & PickPlaceCounterToDrawer    & 2 & 3 & 0 & 1 & 0 & 3 & 2 \\
    LowerHeat            & 3 & 3 & 0 & 0 & 0 & 0 & 1 & CoffeeSetupMug              & 2 & 1 & 0 & 0 & 0 & 1 & 1 \\
    \bottomrule
  \end{tabular}}
\end{table}

\section{Baselines}
\label{app:baselines}

\textbf{Common interface.}
Both code-writing baselines call the functions of \cref{tab:capapi}, modelled on CaP-X's low-level interface \citep{fu2026capx}: names, arguments, return formats and docstrings follow its Franka interface where the robot allows, so that CaP-X's prompts, which are assembled from the docstrings, carry over.
Every pose in our docstrings, including the target of \texttt{solve\_ik}, refers to the point between the fingertips.
Segmentation is SAM~3 and grasps come from Contact-GraspNet, both run by the services of CaP-X and called with its request formats. The VLM of the episode points to image locations in place of Molmo, which we did not serve.
Because the robot is a mobile manipulator, \texttt{solve\_ik} and \texttt{move\_to\_joints} move the arm with the base in place, as on the fixed-base arm of CaP-X. \texttt{goto\_pose}, which the high-level and LIBERO interfaces of CaP-X also offer, places the base and torso, plans a collision-free arm path among the surfaces that the depth cameras saw and approaches the pose straight along the axis of the gripper, using the planner and controller of \method{}. \texttt{navigate\_to\_pose} moves the base.
Neither original has a function for articulated parts, and neither has one here: a program computes arcs and slides itself.
Both baselines get the starting images of \method{} (three head cameras and the upward wrist view) and the same notes on the tasks' conventions, rewritten without references to \method{}'s own tools.

\begin{table}[h]
  \caption{Functions of the code-writing baselines. CaP-Agent0 gets the low-level set and CaP-X's synthesized skill library (as in the M4 tier of CaP-X with the library), and Code as Policies gets the high-level functions of CaP-X, built from the same pieces.}
  \label{tab:capapi}
  \centering
  \small
  \begin{tabular}{@{}p{3.1cm}p{11.2cm}@{}}
    \toprule
    set & functions \\
    \midrule
    perception & \texttt{get\_observation} (RGB-D, intrinsics and pose of each camera), \texttt{segment\_sam3\_text\_prompt}, \texttt{segment\_sam3\_point\_prompt}, \texttt{point\_prompt\_molmo} (answered by the VLM), \texttt{get\_oriented\_bounding\_box\_from\_3d\_points}, \texttt{plan\_grasp} (Contact-GraspNet), \texttt{get\_robot\_state} \\
    motion & \texttt{solve\_ik}, \texttt{move\_to\_joints}, \texttt{goto\_pose}, \texttt{navigate\_to\_pose}, \texttt{goto\_home\_joint\_position}, \texttt{open\_gripper}, \texttt{close\_gripper} \\
    skill library (CaP-Agent0) & \texttt{rotation\_matrix\_to\_quaternion}, \texttt{decompose\_transform}, \texttt{depth\_to\_point\_cloud}, \texttt{mask\_to\_world\_points}, \texttt{pixel\_to\_world\_point}, \texttt{transform\_points}, \texttt{interpolate\_segment}, \texttt{normalize\_vector}, \texttt{select\_top\_down\_grasp} \\
    high level (CaP) & \texttt{get\_object\_pose}, \texttt{get\_object\_poses}, \texttt{sample\_grasp\_pose}, \texttt{goto\_pose}, \texttt{move\_to\_pose}, \texttt{open\_gripper}, \texttt{close\_gripper}, \texttt{get\_robot\_state}, \texttt{goto\_home\_joint\_position} \\
    \bottomrule
  \end{tabular}
\end{table}

\textbf{Code as Policies.}
We use the released LMP code (\texttt{LMP}, \texttt{LMPFGen}, \texttt{FunctionParser}, \texttt{exec\_safe}) with only the model call changed.
A top-level LMP completes a Python script of kitchen examples for the instruction, and a \texttt{parse\_obj\_name} LMP chooses among detected instances by a spatial description. Functions that the program calls but that no one defined are written recursively by the function-generating LMP before the program runs.
The program runs once, with no execution feedback and no image: the VLM sees the scene only through the values that the perception functions return, as in the original.
The examples are new, since CaP has no kitchen or articulated-object prompt, and none is one of the evaluated tasks (a dishwasher door, a fridge door, a toaster lever, fruit into a bowl, a cup chosen by position).
The chat model is told to write only the code that follows the prompt, and the stop sequences of CaP are applied to its answer. A reasoning model takes no temperature, so it does not decode greedily as the original did.

\textbf{CaP-Agent0.}
We use the released agent loop and prompts. The coder writes a whole program, which runs in a persistent namespace. It then receives the executed code, the console output and errors, and a description by the visual differencing module of how the scene changed and whether the task looks complete, and it answers REGENERATE with new code, or FINISH, for at most ten regenerations.
The differencing module also describes the initial scene.
Each query draws three candidates, which a synthesizer merges, as in the released configuration for OpenAI models. The coder never sees an image, and we use the skill library of nine functions verbatim.
The differencing module sees the three head cameras and the upward wrist view tiled in one image, where CaP-X uses one camera.

\textbf{ReKep.}
We run the released code with our robot and scenes in place of OmniGibson: its keypoint proposal (DINOv2 features clustered within each mask), its constraint-generation prompt, its sub-goal and path solvers and its re-planning loop with backtracking.
As in the release, masks come from the segmentation of the simulator (here one per rigid part), and the keypoints follow the true poses of their parts.
We take the image from the overview pose of the other methods, after which the base returns to its start pose and stays there, as in the release.
Inverse kinematics solves for the seven arm joints.
We made three changes to keep the released loop running in kitchens.
First, the obstacle field takes the sign of each distance from a majority of ray-crossing parities.
The signed distance of open3d, which the release uses, puts free space next to the coincident faces of RoboCasa fixtures inside obstacles, including the start pose of the gripper.
Second, each re-plan executes 20 waypoints instead of 5, as the warning of the released loop advises when the robot stops moving.
Third, inverse kinematics uses the same arithmetic in a faster implementation.
Grasping is physical, as for every method, whereas the release attaches a grasped object to the gripper. Only 13 of the 2,593 grasp attempts of ReKep hold.
Episodes have the time limit of the other methods.
The released parser cuts a constraint function at a \texttt{return} statement that spans several lines, which GPT-6.1-Sol often writes, and so ends 13 episodes before any motion.
In one more episode, it finds no number of stages in the answer.
With Python's own parser, which runs all 13, ReKep succeeds in 6 episodes instead of 5.

\textbf{Model.}
Every role of the three zero-shot baselines uses GPT-6.1-Sol at medium reasoning effort, as \method{} does, through the Codex CLI with its tools disabled, so that only our harness executes code.
Every query starts a new conversation.

\textbf{GR00T N1.5.}
The released RoboCasa365 pretraining checkpoint runs with 16-step action chunks, three cameras at $256 \times 256$ and the task's horizon times 1.5.
Its evaluation code counts an episode as solved if the success check held at any step. In the same run, this protocol gives 52 successes, compared with 42 at the end of the episode.
The multitask checkpoint scored 44 of 90 by that protocol.

\textbf{$\pi_{0.5}$.}
We use the RoboCasa365 multitask checkpoint of $\pi_{0.5}$, trained on human demonstrations of 300 tasks that include all 30 of ours.
It runs from the openpi fork of the RoboCasa team with the protocol of GR00T, which sets the cameras and actions of the gym wrapper and 1.5 times the horizon of the task.
Scored at the end of the episode it solves 35 of 90, and 36 if success at any step counted.
On LIBERO-PRO we run the \texttt{pi05\_libero} checkpoint with the LIBERO client of openpi and the quantile normalization it was trained with, taking the instruction from the task file of each episode.

\textbf{Multi-turn coding agent.}
Before the published baselines, we ran a multi-turn coding agent of our own in the style of CodeAct \citep{wang2024codeact}, with the newer GPT-6.1-Sol. In each turn, the VLM returns its reasoning, a decision (run, done or fail) and code, which runs in a persistent namespace against the live simulator. The interface gives head and upward images, depth points with fitted normals, the mapped point cloud and plane fits, crops, wrist and label views, and the hand pose, and for motion \texttt{reach} (base and torso placement, a collision-free arm path and a guarded straight approach), \texttt{grip}, a force-guarded compliant \texttt{move}, a compliant \texttt{swing} about a hinge, and \texttt{release}. Printed output, errors and requested images return in the next prompt, for up to 30 turns.
It succeeded in 61 of the 90 kitchen episodes with 18.6 calls and 0.59M input tokens per episode (9 false successes). It ran before a fix that made RoboCasa's placement of objects beside sinks and stoves deterministic: in four episodes its distractor objects sit on the other side of the sink and one is unresolved, and it failed all five.

\textbf{Further baselines with an earlier version.}
With an earlier version of \method{} we also compared three zero-shot designs that share its VLM, robot and executor but have no world model, hypotheses or verifier, on 50 episodes of the 15 common tasks with other seeds (\cref{tab:baselines_old}): \emph{direct poses} (the VLM writes grasp poses and waypoints), a \emph{skill orchestrator} (the VLM calls a fixed library of open, close, turn, press and pick-and-place skills) and \emph{keypoint constraints} (the VLM writes constraints on tracked keypoints that an optimizer solves), plus the multi-turn coding agent with GPT-6-Astra.

\begin{table}[h]
  \caption{Further baselines on 50 episodes of the 15 common tasks, with an earlier version of \method{} (two runs, mean) and one partial run per baseline, all with GPT-6-Astra. Unfinished arms are compared with ours on the episodes they finished. $\Delta$: difference with a bootstrap 95\% interval.}
  \label{tab:baselines_old}
  \centering
  \small
  \begin{tabular}{lrrrrrr}
    \toprule
    method & $n$ & success & ours, same & $\Delta$ (95\% CI) & calls & input tokens \\
    \midrule
    \method{} (earlier version) & 50 & 38.5 & --   & --                & 4.2  & 0.13M \\
    direct poses             & 47 & 28   & 37   & $-9.0$ ($-15.5$, $-3.0$) & 7.3  & 0.23M \\
    skill orchestrator       & 50 & 26   & 38.5 & $-12.5$ ($-19.0$, $-6.5$) & 6.9  & 0.21M \\
    keypoint constraints     & 26 & 16   & 21.5 & $-5.5$ ($-10.5$, $-0.5$) & 7.4  & 0.35M \\
    multi-turn coding agent  & 21 & 14   & 16.5 & $-2.5$ ($-8.0$, 3.5)  & 17.9 & 0.58M \\
    \bottomrule
  \end{tabular}
\end{table}

\textbf{Coding agent on LIBERO-PRO.}
On LIBERO-PRO, our coding agent (GPT-6.1-Sol) writes Python against an RGB-D and point-cloud interface (captures, clusters, a top-down map, sharp zooms) with the same planner and controller as \method{} (\texttt{move\_to}, \texttt{pick}, \texttt{place}, \texttt{push}), for up to 20 turns. The scene is static while it thinks, and it never sees object poses or a success signal.
Its prompt was developed on the original LIBERO suites at another initial state.

\textbf{CaP-X on LIBERO-PRO.}
We run the code and prompts of CaP-X unchanged on our 150 episodes, with its LIBERO configuration of CaP-Agent0.
This configuration uses the reduced functions with the 9-function skill library, up to ten regenerations, three candidates and a synthesizer per query, and visual differencing.
GPT-6.1-Sol answers every model call, including the pointing requests that CaP-X sends to Molmo.
An episode may take 8,000 simulation steps, and CaP-X restarts a trial that takes longer than 1000\,s, for up to three attempts.
Since our calls share three VLM connections, we charge the response time of each call to the trial but not its wait for a free connection.
The single-turn tier with low-level functions and examples (S3) uses the same configuration without multi-turn prompting, visual differencing or the ensemble.
Each episode starts from the released initial state of LIBERO-PRO, which the reset of CaP-X would otherwise re-sample.
The success rates per task that CaP-X reports \citep[App.~J]{fu2026capx} predict 30 successes of CaP-Agent0 on our episodes (95\% range 22--39).
It succeeds in 23.
Of its 127 failed episodes, 78 run out of the 8,000 steps, 48 use all eleven programs and one ends with a false claim of success.
The largest shortfall is on turning on the stove with perturbed positions, which it fails in all five episodes, where the reported rate predicts 3.3 successes.

The interface of CaP-X documents \texttt{solve\_ik} as solving for the \texttt{panda\_hand} link, and the pose it reports for the robot as that of \texttt{panda\_hand}.
The documentation of \texttt{plan\_grasp} leaves the offset from the hand to the fingertips to the caller.
However, \texttt{solve\_ik} places the hand 10\,cm behind the target it is given, and the oracle programs of CaP-X for LIBERO pass a grasp position to it unchanged.
The first program applies the offset itself (10.34\,cm for the Franka gripper) in all 150 single-turn episodes and in 126 of the CaP-Agent0 episodes.
These include 22 of its 23 successes, whose later programs drop the offset or measure and compensate it.
In the failed episodes of CaP-Agent0, half of the programs stop at checks that the agent wrote itself.
About a third of its motion checks compare the commanded target with the reported pose of the hand, which a correct motion leaves 11\,cm apart.
Of the 47 motion checks whose program printed this gap, 31 show 9--13\,cm.

\section{Requesting versus Selecting}
\label{app:early_minimal}

In \emph{selection-based} construction, a geometric backend proposes everything and the VLM selects and annotates. The backend segments all four images with the automatic mask generator of SAM~2 \citep{ravi2025sam2}, fits a primitive to every mask and draws the candidates as numbered marks \citep{yang2023setofmark}. The VLM then picks the task-relevant candidates, gives them roles (link, part) and adds the joint, hypotheses and goal as in an assembly.
We compared it with requests on 50 episodes with the same sensors, VLM, planner, executor and verifier (\cref{tab:minimal}). In the requests arm, the VLM asked for masks of the parts it needed, segmented by the same model.
Two further arms bound the design: \emph{coarse} keeps only object-level boxes and planes, and \emph{full} requests like ours but models every articulated part it sees, not only the target.
Requests extract 190 times fewer primitives than the selection inventory and succeed in 8 more of the 31 episodes both finished ($p = 0.02$), and grounding the requests in points instead of masks raised success further, to 36 of the 50 episodes.

\begin{table}[h]
  \caption{Requesting versus selecting from an inventory, with an earlier version of the system (same segmenter, planner, executor and verifier; 50 episodes, one run each; unfinished arms are compared on shared episodes). \emph{Requests}: the VLM asks for masks of the parts it needs; \emph{selection}: the backend segments and fits everything and the VLM selects; \emph{coarse}: object-level boxes and planes only; \emph{full}: requests, but every visible articulated part is modelled.}
  \label{tab:minimal}
  \centering
  \small
  \setlength{\tabcolsep}{3.4pt}
  \begin{tabular}{lcccc}
    \toprule
     & requests & selection & coarse & full \\
    \midrule
    successes / finished       & 26/40 & 14/34 & 4/50 & 11/22 \\
    right part bound           & 18/40 & 17/27 & 6/27 & 8/21 \\
    $\Delta$ vs requests, shared & --  & $-8$/31 & $-22$/40 & $-4$/22 \\
    McNemar $p$                & --    & 0.02  & $<10^{-5}$ & 0.22 \\
    primitives extracted       & 7     & 1361  & 91   & 118 \\
    primitives in model        & 2.8   & 3.0   & 1.8  & 52 \\
    VLM calls                  & 5.1   & 1.9   & 1.9  & 6.8 \\
    construction (s)           & 89    & 92    & 47   & 433 \\
    \bottomrule
  \end{tabular}
\end{table}

\begin{table}[h]
  \caption{Task-directed requests against selection from an inventory with an early version of the system and executor (9 tasks $\times$ 5 seeds; medians unless noted). \emph{Binding}: the bound part lies within a few centimetres of the true handle, knob, lever or button (30 fixture episodes). Task success was low for all three methods with the executor of that time and does not separate them.}
  \label{tab:minimal_early}
  \centering
  \small
  \setlength{\tabcolsep}{3.5pt}
  \begin{tabular}{lccc}
    \toprule
     & requests & exhaustive & coarse \\
    \midrule
    primitives extracted       & 7     & 1213  & 94   \\
    primitives in model (mean) & 2.1   & 10.9  & 3.4  \\
    correct binding            & 25/30 & 17/30 & 3/30 \\
    binding error              & 1.0\,cm & 3.1\,cm & 25\,cm \\
    construction time          & 70\,s & 97\,s & 52\,s \\
    uncached input and output tokens & 55k & 79k & 22k  \\
    task success               & 11/45 & 13/45 & 5/45 \\
    \bottomrule
  \end{tabular}
\end{table}

Requests extracted about 170 times fewer primitives and bound the right part more often, most clearly for small parts (buttons 4/5 compared with 1/5, faucet levers 5/5 compared with 1/5).
In the comparison of \cref{tab:minimal}, selection bound the right part at least as often as requests (17 of 27 episodes, compared with 18 of 40), yet its episodes failed more often, with no feasible plan in 7 of 20 failures. We have not isolated the cause.

\section{Grounding by Points and by Masks}
\label{app:grounding}

Grounding by lifted points builds more correct models than grounding by masks on 30 episodes (23 compared with 15, with the truth among the hypotheses in 27 compared with 15), in less time (49 compared with 62\,s) and with the same three VLM calls.
The gain comes from small parts and senses: a knob or a button is one snapped point, a lever two points, and hypotheses carry what the images do not show.
Masks are better for doors: a mask's measured surface places a door's hinge edge within about 2\,cm, while hinge lines from pointed corners are off by about 4\,cm at a distance of 1.5\,m.
In 9 of these 30 episodes the VLM corrected its own model after seeing the overlay.
\Cref{tab:variants} compares the three groundings on 50 further episodes with an earlier version.

\begin{table}[h]
  \caption{Three groundings on 50 episodes of the 15 common tasks (other seeds) with an earlier version, successes summed over two runs (of $2n$). \emph{Union}: solved by any grounding in the same run.}
  \label{tab:variants}
  \centering
  \small
  \begin{tabular}{lrcccc}
    \toprule
    family & $n$ & points & hybrid & masks & union \\
    \midrule
    drawers       &  8 & 12 & 14 & 14 & 14 \\
    doors         & 20 & 28 & 23 & 21 & 30 \\
    knobs         &  6 & 10 &  9 &  7 & 10 \\
    buttons       &  6 & 12 & 11 & 11 & 12 \\
    levers, spout &  8 & 12 & 15 & 12 & 16 \\
    pick \& place &  2 &  3 &  3 &  0 &  4 \\
    \midrule
    all           & 50 & 77\% & 75\% & 65\% & 86\% \\
    model correct &    & 71\% & 75\% & 46\% & \\
    \bottomrule
  \end{tabular}
\end{table}

\section{Component Ablations}
\label{app:repeats}

\begin{table}[h]
  \caption{Component ablations of an earlier version on 60 episodes of the 15 common tasks (seeds 120--179), one run per arm (the full system: three runs).}
  \label{tab:ablate}
  \centering
  \small
  \begin{tabular}{lcc}
    \toprule
    arm & success & model correct as built \\
    \midrule
    full system (3 runs)    & 78\% (141/180) & 69\% \\
    no verifier             & 81\% (48/59)   & -- \\
    no requested views      & 80\% (47/59)   & 71\% \\
    single hypothesis       & 78\% (46/59)   & 66\% \\
    no overlay check        & 76\% (45/59)   & 59\% \\
    \bottomrule
  \end{tabular}
\end{table}

With one run per arm the intervals are about $\pm 9$ points, so none of these removals shows a success effect at this size.

With the final system, 11 of the 82 successes of the first run with GPT-6-Astra came only after a restart.
On the 35 episodes of a run whose success rested on weak evidence, the final check kept all 26 true successes and turned 2 of the 9 false successes into successes.

\section{Other VLMs}
\label{app:vlms}

\begin{table}[h]
  \caption{Other VLMs, with earlier versions of \method{}. Top: 60 episodes of the 15 common tasks (runs summed). Bottom: 30 of the kitchen episodes (the first seed of each task), with a version that solved 53 of the 90, and the multi-turn coding agent of \cref{app:baselines} on the same episodes.}
  \label{tab:vlms}
  \centering
  \small
  \resizebox{\textwidth}{!}{\begin{tabular}{llccl}
    \toprule
    episodes & VLM & \method{} & coding agent & note \\
    \midrule
    \multirow{3}{*}{60} & GPT-6-Astra (default) & 78\% of 180 & -- & \\
     & Claude Sonnet 5.5 & 61\% of 90 & -- & models correct as built 38\% against 69\% \\
     & Qwen3.5-9B (open weight) & 2\% of 57 & -- & copies the prompt's example coordinates \\
    \midrule
    \multirow{3}{*}{30} & GPT-6 (Astra; Sol for the agent) & 19 & 22 & first sketch correct 23/30 \\
     & Gemini 4 Argon & 17 & 24 & first sketch correct 18/30; paired $p = 0.23$ \\
     & Qwen3.6-35B (open weight) & 1 & 0 & \\
    \bottomrule
  \end{tabular}}
\end{table}

With Claude, levers succeed in 4 of 16 runs (23 of 30 for GPT-6-Astra) and doors in 20 of 40 (44 of 66).
One command-line harness per vendor was used, and token counts are not comparable across vendors.
A run with a further open-weight model through an API service had infrastructure failures in most episodes and is not reported.

\section{Perception Robustness}
\label{app:robust}

\begin{table}[h]
  \caption{Perception noise with an earlier version: the VLM's outputs are replayed from a baseline run and only the perception is degraded (30-episode subset, one run per condition, 23--30 episodes each). Clean: the same episodes without noise.}
  \label{tab:robust}
  \centering
  \small
  \begin{tabular}{lcc}
    \toprule
    condition & success & clean, same episodes \\
    \midrule
    clean replay                          & 20/30 & -- \\
    depth noise $\sigma = 0.5$\,cm        & 17/28 & 19 \\
    depth noise $\sigma = 1$\,cm          & 9/23  & 16 \\
    depth noise $\sigma = 2$\,cm          & 4/30  & 20 \\
    range-proportional noise 2\%          & 1/29  & 20 \\
    10\% of pixels dropped                & 21/30 & 20 \\
    six patches removed                   & 11/28 & 18 \\
    extrinsic error 1\,cm / $1^\circ$     & 14/24 & 18 \\
    extrinsic error 2\,cm / $2^\circ$     & 7/28  & 20 \\
    combined mild                         & 2/30  & 20 \\
    \bottomrule
  \end{tabular}
\end{table}

The correct models fall from 67\% to 35\% at 1\,cm of depth noise and 23\% at 2\,cm: the loss comes from wrong models, not from failed execution.
Replay cannot re-point the camera after a missing depth reading, so the patch results are pessimistic. The noise is simulated, not measured on a sensor.
On LIBERO, a scripted version of the tabletop pipeline loses most of its success at 1\,cm of depth noise (75\% to 28\% of 40 episodes), and a $5 \times 5$ median filter on depth keeps 68\% at that level.

\section{Composite Tasks}
\label{app:composite}

With an earlier version we ran 10 composite tasks of two to four steps (30 paired episodes), with the step list given.
Keeping one sketch for the whole episode and adding only the next step's parts takes 7.8 VLM calls per episode, compared with 10.2 when each step is sketched from scratch (paired difference 2.3 calls, interval 1.6 to 3.1), and 2.9 compared with 4.4 calls in the later steps.
The task succeeds in 3 of 30 episodes with the persistent sketch and 5 of 30 without (not significant): the second step of most tasks (placing an object, pressing a start button, a small lever) already failed on its own at the time.

\section{Navigation Details}
\label{app:nav}

\textbf{Setting.}
We follow the R2R-CE task of VLN-CE \citep{krantz2020vlnce}: forward steps of 0.25\,m and turns of $15^\circ$ with sliding along walls, at most 500 actions, success when the robot stops within 3\,m (geodesic) of the goal, one camera 1.25\,m above the floor with a $90^\circ$ horizontal field of view, no actuation or depth noise, and no GPS or compass.
Three settings differ.
Our RGB-D images have 640$\times$480 pixels, where VLN-CE uses 224$\times$224 RGB and 256$\times$256 depth images.
Depth is clipped to 0.5--5\,m, and the robot takes a panorama by turning in place, which costs 20 of its 500 actions.
We compute the metrics as VLN-CE does, with oracle success and the final distance from geodesic distances after every action, SPL from the 3-D path length, and nDTW with fastdtw on the dense reference path.

\textbf{Agent.}
Each episode is one VLM conversation, resumed at each turn, with a fresh conversation and a written summary after 8 turns and at most 20 turns in all.
A turn receives a panorama, or only the forward view if the VLM asks for it, and an overhead map built from depth.
In a turn the VLM can look in a chosen direction, lift pixels to 3-D points, edit the sketch and send the robot towards a primitive, possibly as the final goal.
A 3-D least-cost path planner and a discrete controller execute the motion, and depth checks the sketch at every keyframe.
The robot estimates its pose by dead reckoning on the commanded actions, corrected by registering 160$\times$120 depth images with earlier scans.

\textbf{Episodes.}
The 376 episodes take one instruction for each val-unseen path that we did not use in development, and they cover all 11 scenes.
The 100 Open-Nav episodes come from 10 scenes.
As \citet{zhou2026mip} do, we take them from the v1-3 data of VLN-CE, which matches the Open-Nav release in start positions, goals and instructions but not in start headings.
Thirteen of them were in our development set, and 24 more share a path with a development episode.
\Cref{tab:nav_subsets} gives the results on these subsets.
In 59 of the Open-Nav episodes all three runs succeed, and in 10 none does.
An Open-Nav episode takes 110k input tokens (26k uncached) and 2.4k output tokens on average.

\begin{table}[h]
  \caption{Navigation results by episode subset (GPT-6.1-Sol, medium effort). For the Open-Nav episodes, SR is the mean and s.d.\ over three runs, \emph{outside dev.}\ leaves out the 13 development episodes and \emph{path-disjoint} also the 24 that share a path with one. The unused episodes have one run.}
  \label{tab:nav_subsets}
  \centering
  \small
  \begin{tabular}{llcccccc}
    \toprule
    episodes & & SR (\%) & SPL (\%) & NE (m) & OSR (\%) & nDTW (\%) & VLM calls \\
    \midrule
    Open-Nav & all (100) & 75.7 $\pm$ 1.2 & 57.5 & 2.86 & 83.0 & 67.4 & 5.7 \\
     & outside dev.\ (87) & 74.7 $\pm$ 1.2 & 55.9 & 2.86 & 82.8 & 67.2 & 5.7 \\
     & path-disjoint (63) & 77.8 $\pm$ 2.8 & 57.7 & 2.49 & 83.6 & 67.5 & 5.9 \\
    \midrule
    unused & all (376) & 76.9 & 56.6 & 2.60 & 82.2 & 67.2 & 6.3 \\
     & single-floor (306) & 75.8 & 56.5 & 2.79 & 81.0 & 66.6 & 5.7 \\
     & multi-floor (70) & 81.4 & 57.4 & 1.79 & 87.1 & 69.6 & 9.3 \\
    \bottomrule
  \end{tabular}
\end{table}

The trajectories of successful runs are longer than the shortest path but stay mostly near the reference route.
On the Open-Nav episodes the median successful run travels 1.23 times the geodesic distance to the goal.
When we project each trajectory onto the reference route, 2\% of its length goes back along the route, 22\% weaves within 2\,m of it and 12\% lies further away.
Given the goal and the staircases on the route, the planner and pose estimate reach all 60 goals of a development set with an SPL of 89\%, so most of the extra length comes from the intermediate targets that the VLM chooses.



\end{document}